%% file: main.tex
\documentclass{article}

\usepackage[nonatbib,preprint]{neurips_2021}

\usepackage{mathtools}

\usepackage[utf8]{inputenc} 
\usepackage[T1]{fontenc}    
\usepackage{url}            
\usepackage{booktabs}       
\usepackage{amsfonts}       
\usepackage{nicefrac}       
\usepackage{microtype}      
\usepackage{xcolor}         
\usepackage{dutchcal}
\usepackage{url}
\usepackage[ruled,vlined,linesnumbered,noresetcount]{algorithm2e}
\usepackage{footnote}
\usepackage[english]{babel}
\usepackage{tikz}
\usepackage{multirow}
\usepackage{arydshln}
\usepackage{booktabs}
\usepackage{textcomp}
\usepackage{xspace}
\usepackage{stfloats}
\usepackage{float}
\usepackage{xcolor}
\usepackage{graphicx}
\usepackage{multirow}
\usepackage{subfiles}
\usepackage{subfig}
\usepackage{wrapfig}
\usepackage{multicol}
\usepackage{adjustbox}
\usepackage{framed}
\definecolor{shadecolor}{rgb}{0.88,0.93,0.93}
\usepackage{cleveref} 
\usepackage[utf8]{inputenc}
\usepackage[english]{babel}
\usepackage{amsthm}
\usepackage{fancyhdr}
\newsavebox{\tempbox}
\theoremstyle{definition}
\newtheorem{definition}{Definition}[section]

\newtheorem{proposition}{Proposition}[section]

\newtheorem{remark}{Remark}
\usepackage{environ}         
\usepackage{etoolbox}        
\newlength{\myl}
\let\origequation=\equation
\let\origendequation=\endequation

\RenewEnviron{equation}{
  \settowidth{\myl}{$\BODY$}                       
  \origequation
  \ifdimcomp{\the\linewidth}{>}{\the\myl}
  {\ensuremath{\BODY}}                             
  {\resizebox{\linewidth}{!}{\ensuremath{\BODY}}}  
  \origendequation
}

\title{Is Shapley Value fair? \\
Improving Client Selection for Mavericks in Federated Learning}

\author{%
 Jiyue Huang \\
 Delft University of Technology\\
  \texttt{j.huang-4@tudelft.nl} \\
   \And
   Chi Hong \\
 Delft University of Technology\\
   \texttt{c.hong@tudelft.nl} \\
   \AND
   Lydia Y. Chen$^*$ \\
 Delft University of Technology\\
   \texttt{lydiaychen@ieee.org} \\
   \And
   Stefanie Roos$^*$ \\
 Delft University of Technology\\
   \texttt{s.roos@tudelft.nl} \\
}

\begin{document}

\newcommand{\svb}{\textsc{SVB}\xspace}
\newcommand{\alg}{\textsc{FedEMD}\xspace}
\newcommand{\federator}{\emph{federator}\xspace}
\newcommand{\SV}{\emph{Shapley Value}\xspace}
\newcommand{\iid}{\emph{i.i.d.}\xspace}
\newcommand{\niid}{\emph{non-i.i.d.}\xspace}
\crefname{section}{§}{§§}


\maketitle

\begin{abstract}

\SV is commonly adopted to measure and incentivize client participation in federated learning. In this paper, we show --- theoretically and through simulations--- that \SV  underestimates the contribution of a common type of client: the Maverick.  Mavericks are clients that differ both in data distribution and data quantity and can be the sole owners of certain types of data. Selecting the right clients at the right moment is important for federated learning to reduce convergence times and improve  accuracy. We propose \alg, an adaptive client selection strategy based on the Wasserstein distance between the local and global data distributions. As \alg adapts the selection probability such that Mavericks are preferably selected when the model benefits from improvement on rare classes, it consistently ensures the fast convergence in the presence of different types of Mavericks.   
Compared to existing strategies, including \SV-based ones, \alg improves the convergence of neural network classifiers by at least 26.9\% for FedAvg aggregation compared with the state of the art. 

\end{abstract}
\vspace{-1em}
\input{samples/Sections/introduction}

\vspace{-1em}
\input{samples/Sections/background}
\vspace{-1em}
\input{samples/Sections/shapley}

\vspace{-1em}
\input{samples/Sections/selection}

\vspace{-1em}
\input{samples/Sections/Evaluation}
\vspace{-1em}
\input{samples/Sections/conclusion}

\bibliographystyle{plain}
\bibliography{main}

\clearpage

\appendix
\input{samples/Sections/supplement}

\end{document}

%% file: samples/Sections/introduction.tex
\section{Introduction}
\label{sec:introduction}
Federated Learning (FL)~\cite{yang2019federated,mcmahan:2017:aistat:fedavg,GhoshCYR20Efficient,reisizadeh2020robust} enables privacy-preserving learning by deriving local models and aggregating them into one global model, meaning that sensitive private data, e.g., health records, never leaves the user's personal devices. Thus, Federated Learning allows the construction of models that cannot be computed on a central server and is hence often the only alternative for medical research and other domains with high privacy requirements.

Concretely, a central server, the \federator, selects \emph{clients} that train local models using their own data. The \federator then aggregates these local models into a global model. This process is repeated over several global rounds, typically with different sets of clients selected in each round. 
Consequently, clients have to invest resources --- storage, computation, and communication --- into a FL system.
It is essential for the \federator to measure the contributions of clients' local models. Based on the contribution, the \federator can select the most suitable clients to complete the learning fast and accurately, in addition to awarding rewards based on contribution. 

There exist a number of proposals for contribution measurement, i.e., algorithms that determine the quality of the service provided by the clients, for Federated Learning~\cite{Kang:2019:IEEEIoT:contract,aono:2017:TIFS:gradient,Richardson:2019:IJCAIFL:Rewarding,wang:2019:Bigdata:contribution,Yuan:20:fedcoin,Song2019Profit,wang:2020:FL:FedSV,richardson:2020:FL:budget,wei:2020:FL:efficientSV}.
In particular, previous work established that 
\SV, which measure the marginal loss caused by a client's sequential absence from the training, offer accurate contribution measurements.
Yet, the prior art of \emph{contribution measurement} focused on identical and independently distributed (\iid) data~\cite{wang:2019:Bigdata:contribution}. Here, \iid refers to \iid data distributions whereas the data quantity can be heterogeneous. The assumption of \iid is quite limiting in practice. 
For example, in the widely used image classification benchmark, Cifar-10~\cite{krizhevsky:2009:cifar}, most people can contribute images of cats and dogs.
However, deer images are bound to be comparable rare and owned by few clients. 
Another relevant example arises from learning predictive medicine from clinics who specialize in different kinds of patients, e.g., AIDS and Amyotrophic Lateral Sclerosis, and own data of exclusive disease type. 
\begin{wrapfigure}{r}{7cm}
\centering
\includegraphics[width=0.5\textwidth]{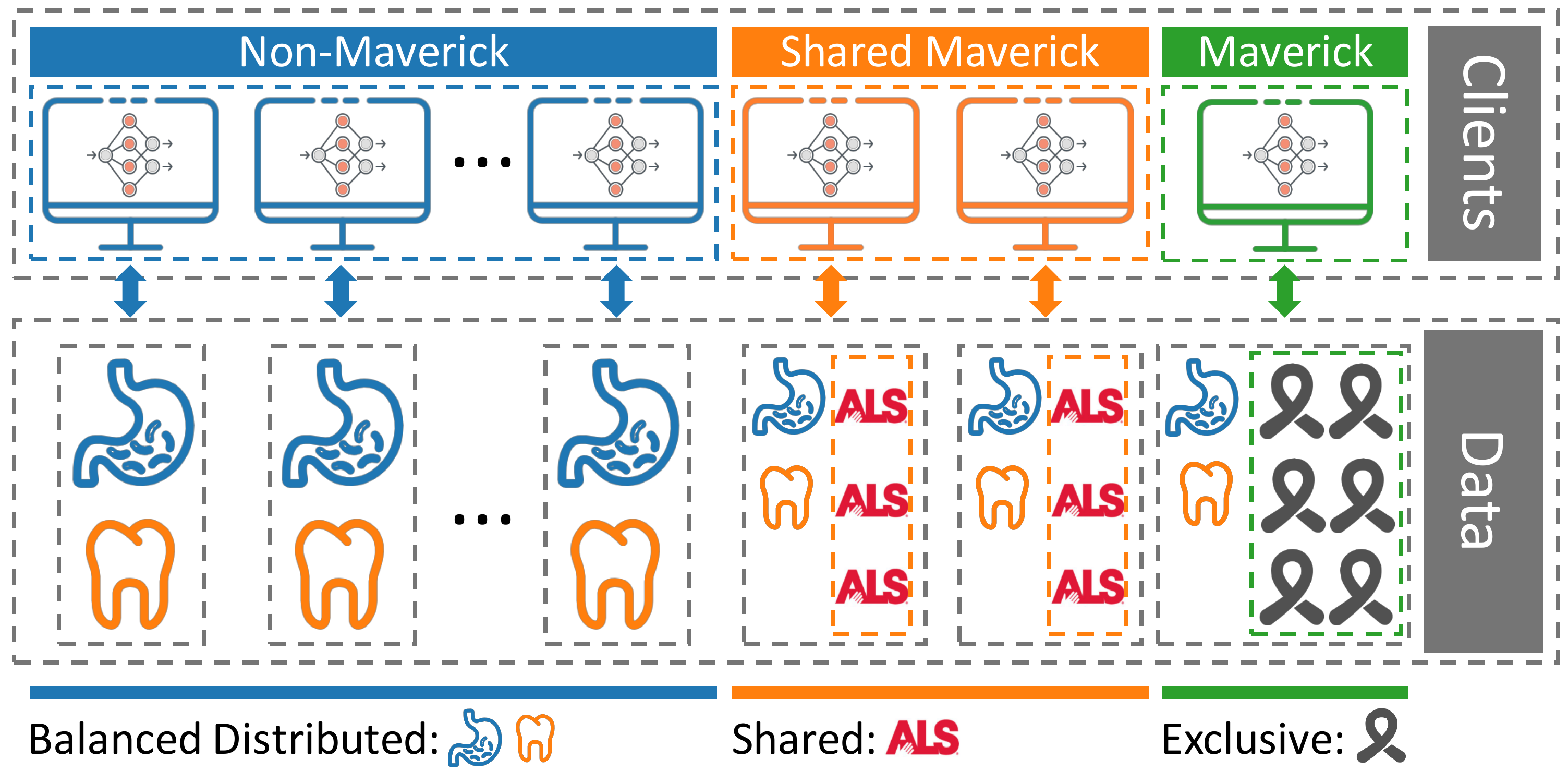}
\setlength{\abovecaptionskip}{0.2cm}
\caption{Illustration of Mavericks. }
\label{img:maverick}
\vspace{-1em}
\end{wrapfigure}
We call such clients, who hold data sets from a majority of a subset of classes, \emph{Mavericks}. 
Fig.~\ref{img:maverick} illustrates the different distributions: In a balanced distribution, all clients own data from all classes. If a system has Mavericks, they own one or more classes (almost) exclusively whereas the non-Maverick clients have a balanced distribution for the remaining classes. Multiple Mavericks could own separate classes or jointly own one class. The latter is referred to as \emph{Shared Mavericks}. 
Mavericks are essential to learn high-quality global models in FL. Without Mavericks, it is impossible to achieve high accuracy on the classes for which they own the majority of all training data. Such classes could be, e.g., rare deceases, which can hence only be successfully classified if Mavericks are included. 

\textbf{Hypothesis on the Impact of  Mavericks}.  %
While \SV is shown to be effective in measuring contribution for the \iid case, it is largely unknown if \SV can assess the contribution of  Mavericks fairly and effectively involve them via the selection strategy. In order to form a hypothesis, we first conduct experiments of learning a neural network classifier (See \ref{supp:hypothesis} in Supplement). 
It shows that random and \SV-based selection both perform well in the scenarios that they are designed for. Yet, they both exhibit slow convergence for Mavericks, with \SV not showing a clear advantage over the random selection. 
This result leads us to conjecture that \SV is unable to fairly evaluate Mavericks' contribution. Moreover, this example calls for a novel client selection strategy that can handle Mavericks appropriately. 



\textbf{Contributions}.
The initial experimental study motivates the two contributions of our paper: \textit{i)} a thorough theoretical analysis that shows that indeed \SV-based contribution measurements underestimate the contribution of Mavericks, and \textit{ii)} \alg, a novel client selection based on the Wasserstein distances of a user's data distribution to both the other users' distributions and the global distribution. 

Concretely, the theoretical analysis proves that Mavericks received low contribution measurement scores during the initial stage of the learning process. Intuitively, the reason for these scores lies in the difference of their data and hence model updates to the global model. In latter stages of the learning process, Mavericks' scores converge towards the scores of other users. Intuitively, the decaying learning rate and the increased focus on small improvements for individual classes explain this increase in score for Mavericks. However, adapting a fairness measure that relates data quantity to contribution score, Mavericks, who tend to own more data than others, are still treated unfairly. 
The theoretical analysis is hence in line with our initial experimental results. More generally, the theoretical analysis reveals that clients with a skewed data distribution or a high data amount are not treated fairly by \svb. 

We propose \alg, a light-weight Maverick-aware selection method for Federated Learning. As Mavericks are strategically involved when they can contribute the most, the convergence speed increases, meaning that the learning process can terminate faster and more efficiently. Our emulation results show that the proposed \alg has faster convergence in comparison to state-of-the-art algorithms~\cite{mcmahan:2017:aistat:fedavg,Zheng:2020:HPDC:Tier,muhammad:2020:KDD:fedfast}, by at least 26.9\% and 11.3\% respectively, under FedAvg and FedSGD aggregations for a range of Maverick scenarios.

%% file: samples/Sections/background.tex
\section{Related Studies}
\label{sec:background}
\vspace{-1em}

\textbf{Contribution Measurement}.
Existing work on contribution measurement can be categorized in two classes:  \textit{i)} local approach: clients exchange the local updates, i.e., model weights or gradients, and measure the contribution of each other, e.g., by creating a reputation system~\cite{Kang:2019:IEEEIoT:contract},  and \textit{ii)} global approach: all clients send all their model updates to the \federator who in turn aggregates and computes the contribution via the marginal loss~\cite{Richardson:2019:IJCAIFL:Rewarding, wang:2019:Bigdata:contribution, Yuan:20:fedcoin,Song2019Profit, wang:2020:FL:FedSV,richardson:2020:FL:budget,wei:2020:FL:efficientSV}. The main drawbacks of local approaches are the excessive communication overhead and the lower privacy due to directly exchanged model updates~\cite{aono:2017:TIFS:gradient}. In contrast, the global approach has lower communication overhead and avoids the privacy leakage to other clients by communicating only with the \federator. Prevailing examples of globally measuring contribution are Influence~\cite{Richardson:2019:IJCAIFL:Rewarding, richardson:2020:FL:budget} and \SV \cite{wang:2019:Bigdata:contribution,wei:2020:FL:efficientSV,wang:2020:FL:FedSV,sim2020collaborative}.  The prior art demonstrates that \SV can effectively  measure the client's contribution for the case when clients' data is \iid or of biased quantity~\cite{sim2020collaborative}. Yet, a recent experimental study\cite{zhang:2020:corr:hierarchically}  demonstrates that the correlation between a user's data quality and its \SV is limited. The results raise doubts whether \SV is really a suitable choice for contribution measurement. However, there is no rigorous analysis on whether \SV can effectively evaluate the contribution from heterogeneous users with skewed data distributions. 

\textbf{Client Selection}.
Selecting clients within a heterogeneous group of potential clients is key to enabling fast and accurate learning based on high data quality. 
The state-of-the-art client selection strategies focus on the resource heterogeneity~\cite{nishio:2019:ICC:FedCS,xu:2020:TWC:long-term,huang:2020:tpds:RBCS-F} or data heterogeneity~\cite{Zheng:2020:HPDC:Tier,cho:2020:corr:convergence,Tian:2020:mlsys:Heterogeneous,chai:2019:usenix:heterogeneity}.
In the case of data heterogeneity, which is a focus of our work, selection strategies~\cite{cho:2020:corr:convergence,goetz:2019:activesampling,Zheng:2020:HPDC:Tier} gain insights on the distribution of clients' data and then select them in specific manners. Goetz et. al~\cite{goetz:2019:activesampling} apply active sampling and Cho et. al \cite{cho:2020:corr:convergence} use  Power-of-Choice to favor clients with higher local loss. 
TiFL \cite{Zheng:2020:HPDC:Tier} considers both resource and data heterogeneity to mitigate the impact of straggler and skewed distribution. TiFL applies a contribution-based client selection by evaluating the accuracy of selected participants each round and choose clients of lower accuracy. FedFast \cite{muhammad:2020:KDD:fedfast} chooses classes based on clustering and achieves fast convergence for recommendation systems. However, there is no selection strategy that addresses the Maverick scenario. 

\textbf{Data Heterogeneity}.
As an alternative to client selection strategies, multiple methodologies have been suggested to properly account for data heterogeneity in FL systems~\cite{li2021noniid,deng2020Distributionally,DinhTN20Personalized,Fallah20Personalized,HanzelyHHR20Lower}. Early solutions require the \federator to distribute a shared global training set~\cite{zhao:2018:corr:noniid}, which is demanding and violates data privacy. Later studies either focus on the local learning stage~\cite{Tian:2020:mlsys:Heterogeneous, karimireddy2020scaffold} or improved aggregation~\cite{wang2020tfednova}. For instance,  FedProx~\cite{Tian:2020:mlsys:Heterogeneous} improves the local objective by adding an additional $L_2$ regularization term, whereas FedNova~\cite{wang2020tfednova} first normalizes the local model updates based on the number of their local steps and aggregates the local models.  
The downside of aforementioned solutions is the requirement of additional computation on clients.  

%% file: samples/Sections/shapley.tex
\section{Fairness Analysis for Shapley Value}
\label{sec:shapley}
The objective is to rigorously and analytically answer the question if \SV is a fair contribution measurement for Mavericks. First, we formalize FL, our assumptions, Maverick, and fairness. Since the main building block of \SV is the Influence Index~\cite{Richardson:2019:IJCAIFL:Rewarding}, we then derive the fairness of Influence Index. Further, we derive the fairness of \SV from the Influence Index and show that \SV-based contribution measurements underestimate the contribution of Mavericks but more generally the contribution of other clients with skewed data distributions or large amounts of data. We empirically verify the lack of fairness for Mavericks in existing contribution measurements.

\textbf{Notation: }
In this paper, we focus on classification tasks.  We denote the set of possible inputs as $\mathcal{X}$ and the set of $C$ class labels that can be associated with the inputs as $\mathcal{Y}=\{ Y_1, Y_2, ... , Y_C\}$. 
In agreement with other works, we let $f\colon \mathcal{X} \xrightarrow{} \mathcal{P}$ assign a probability distribution of potential classes to inputs.
With $\omega$ denoting the learned weights of the machine learning tasks, the empirical risk is $\mathcal{L}(\boldsymbol{\omega})$ for a SGD-based learning process under cross-entropy loss. Furthermore, the learning objective is then generally defined as:
     $\min \mathcal{L}(\boldsymbol{\omega}) = \min \sum_{c=1}^{C} p(y=c) \mathbb{E}_{\boldsymbol{x} \mid y=c}\left[\log f_{c}(\boldsymbol{x}, \boldsymbol{\omega})\right]$.
 
 We are solving the learning objective in a Federated Learning scenario. 
 In an FL system, there is a set $\mathcal{C}$ of $N$ clients. Enumerate the $K$ clients selected in a round by $C_1, \ldots, C_K$. 
Each client $C_{k}$ selected in round $t$ computes local updates  $\omega^{k}_{t}$ and the \federator aggregates the results. 
Concretely, with $\eta$ being the learning rate, $C_k$ updates their weights in the t-$th$ global round by
 \begin{equation}
\boldsymbol{\omega}_{t}^k=\boldsymbol{\omega}_{t-1}-\eta  \sum_{c=1}^{C} p^k(y=c) \nabla_{\boldsymbol{\omega}} \mathbb{E}_{\boldsymbol{x} \mid y=c}\left[\log f_{c}(\boldsymbol{x}, \boldsymbol{\omega_{t-1}})\right].
\end{equation}
The most common aggregation method is an average of the client updates, weighted by the amount of data one owns. Let $n^k$ denote the data quantity, then FedAvg aggregation is 
$\boldsymbol{\omega}_{t}=\sum_{k=1}^{K} \frac{n^{k}}{\sum_{k=1}^{K} n^{k}} \boldsymbol{\omega}_{t}^{k}$. 

\textbf{Assumptions: } \textit{i)} All parties are honest and follow the protocol. \textit{ii)} Clients are homogeneous with respect to other resources, such as computation and communication.  \textit{iii)} The network is reliable with all messages being delivered within a maximal delay. \textit{iv)} The global distribution has high similarity with the real-world (test dataset) distribution. 


\begin{definition}[Maverick]
Let $Y_{Mav}$ be a class label that is primarily owned by Mavericks. In the extreme case, there is one Maverick but it might also be a small set of Mavericks jointly owning the same class. For a client $C_k$, let $q^{Mav}_{k}$ be the fraction of $C_k$'s data that has label $Y_{Mav}$. Then: 
\begin{equation}\label{eq:maverick}
   q^{Mav}_{k} \approx \begin{cases}
   1, \text{if } C_k \text{ is a Maverick} \\
   0, \text{if } C_k \text{ is not a Maverick}.
   \end{cases}
\end{equation}
\end{definition} 

\begin{remark}
\textit {A system might contain more than one Maverick. We refer to Mavericks that own exclusive classes as \emph{exclusive} Mavericks. If the Mavericks all own the same class, we refer to them as \emph{shared} Mavericks. }
\end{remark}

The relation between being a Maverick and the amount of owned data can be essential. Learning success on a class relates closely to the amount of data available for that class. Thus, for the Maverick to have a positive contribution when they are the only one owning a class, they should own more data than others. Our evaluation hence focuses on Mavericks that own considerably more data than other clients, e.g., more than half of the data of all selected clients.
However, the theoretical results regarding fairness hold as long as the Maverick owns at least the average amount of data. 

\begin{definition}[Data Size Fairness]
Denote the contribution measured for client $C_k$ in round $t$ as $c_k(t)$, their relative contribution ratio as $rc_k(t)= c_k(t)/\sum_{i=1}^K c_i(t)$, and 
their local data quantity ratio as $q_k(t) = n^k/\sum_{i=1}^{K} n^{i}$.  
We define a system as fair for $C_k$ in round $t$ if the difference $|q_k(t)-rc_k(t)|$ is small. The concrete fairness of client $C_k$ at round $t$ is  
\begin{equation}\label{eq:fairnessClient}
    U_k(t) = 1 - |{q_k(t)-rc_k(t)}|. 
\end{equation}
\end{definition}

\begin{definition}[Fairness Utility]\label{def:utility}
Fairness utility on a system level:
\begin{equation}\label{eq:fairness}
    U = 1 - \frac{1}{T\cdot K}\sum_{t=0}^{T}\sum_{k=1}^K|{q_k(t)-rc_k(t)}|, 
\end{equation}
\end{definition}

\begin{remark}
\textit{ From Def.~\ref{def:utility}, $U=1$ represents the ideal fairness. }
\end{remark}

Our aim is to show that \SV-based contribution measures achieve a low fairness if Mavericks are involved.
Concretely, we show that the value of Eq.~\ref{eq:fairnessClient} is lower if $C_k$ is a Maverick than if they are a non-Maverick. 

\subsection{Influence Index in FL}\label{subsec:Influence}
In this part, we discuss the fairness of Influence Index for clients. First, we consider the effect of data quantity on fairness, particularly for clients with a very low or high quantity of data.  Second, we analyze the effect of skewed data distributions. 

We use Influence Index as defined by Richardson et.\ al~\cite{Richardson:2019:IJCAIFL:Rewarding}: 
Let $\boldsymbol{\omega}_{t_{/k}}$ denote the weights at round $t$ if $C_k$ is excluded from the aggregation and $\boldsymbol{\omega}_{t}^{i}$ refer to the local updates of $C_i$. 
Then, the Influence Index of $C_k$ in global round $t$ is: 
\begin{equation}\label{eq:inf}
\begin{aligned} 
        Inf(C_k) = \left(\mathcal{L}(\boldsymbol{\omega}_{t} )- \mathcal{L}(\boldsymbol{\omega}_{t_{/k}})\right) = \left(\mathcal{L}(\boldsymbol{\omega}_{t} )- \mathcal{L}\left(\frac{\sum_{i=1}^{K} n^{i}\boldsymbol{\omega}_{t}^i- n^k \boldsymbol{\omega}_{t}^{k}}{\sum_{i=1}^{k-1} n^{i}+\sum_{i=k+1}^{K} n^{i}}\right) \right).
\end{aligned}
\end{equation}

\textbf{Data Size Matters: }
As stated above, Mavericks are likely to own a different amount of data than non-Mavericks. Thus, we first look into how the amount of data affects fairness regardless of the data distribution over classes. 
We relate the Influence Index from Eq.~\ref{eq:inf} to the local data quantity by considering the difference in Influence Index of two clients.  Without loss of generality, let one of the clients be $C_1$ and assume that $C_1$ is not a Maverick, resulting in:

\begin{equation}
\label{eq:infgeneral}
    \begin{aligned}
    & Inf(C_k) - Inf(C_1) =   
     \mathcal{L}\left(\frac{\sum_{i=1}^{K} n^{i}\boldsymbol{\omega}_{t}^{i}- n^1 \boldsymbol{\omega}_{t}^{1}}{\sum_{i=2}^{K} n^{i}}\right) - \mathcal{L}\left(\frac{\sum_{i=1}^{K} n^{i}\boldsymbol{\omega}_{t}^{i}- n^k \boldsymbol{\omega}_{t}^{k}}{\sum_{i=1}^{k-1} n^{i}+\sum_{i=k+1}^{K} n^{i}}\right). 
    \end{aligned}             
\end{equation}



\textbf{\textit{Extreme case: $q_k(t)$ is very large}}:
If $q_k(t) \geq 0.5$, i.e., $n^k \geq 0.5 \sum_{i=1}^{K} n^{i}$,  it follows from Eq.~\ref{eq:inf} that
\begin{equation}
  \mathcal{L}\left(\frac{\sum_{i=1}^{K} n^{i}\boldsymbol{\omega}_{t}^{i}- n^k \boldsymbol{\omega}_{t}^{k}}{\sum_{i=1}^{k-1} n^{i}+\sum_{i=k+1}^{K} n^{i}}\right) \approx \mathcal{L}(n^k(\boldsymbol{\omega_t}-\boldsymbol{\omega_t^k})),
\end{equation}
and hence 
    $Inf(C_k)-Inf(C_1) 
    \approx \mathcal{L}(\boldsymbol{\omega}_{t}) - \mathcal{L}(n^k(\boldsymbol{\omega_t}-\boldsymbol{\omega_t^k}))$.
We argue that
it is highly likely that $Inf(C_1) > Inf(C_k)$ or $Inf(C_1) \approx  Inf(C_k)$, i.e., the right-hand side of the equation is negative or close to 0.   
The difference $\boldsymbol{\boldsymbol{\omega}_t-\boldsymbol{\omega}_t^k}$ represents the difference between $C_k$'s weights rather than weights related to the learning process, which can be expected to be high. 

Initially, $\mathcal{L(\boldsymbol{\omega}_t)}$ is likely to be large but still smaller than the completely random $\mathcal{L}(n^k(\boldsymbol{\omega_t}-\boldsymbol{\omega_t^k}))$ with high probability. It follows that $Inf(C_1) \geq Inf(C_k)$.  When $t$ increases, $\mathcal{L(\boldsymbol{\omega}_t)}$ is expected to decrease while $\mathcal{L}(n^k(\boldsymbol{\omega_t}-\boldsymbol{\omega_t^k}))$ stays high,
and hence indeed $Inf(C_1) >  Inf(C_k)$.
As $C_k$ contributes much more data than $C_1$, both scenarios are unfair. 

 \begin{proposition}
     \noindent \textit{ A client $C_k$ with a large data quantity ratio is treated unfairly as their Influence Index is lower than others despite having a higher local data quantity.}
 \end{proposition}

\textbf{Data Distribution Matters:}
Following a similar approach before,
we adapt a more general notion of a skewed distribution than in Eq.~\ref{eq:maverick} to obtain results that are of relevance beyond the case of Mavericks. 
Concretely, we define a skewed data distribution as one that differs from the global data distribution significantly, e.g., more than expected when distributing data randomly between clients. 
In order to analyze the impact of distributions, consider the Kullback-Leibler Divergence (KLD)~\cite{kullback:1951:MS:KLD}, which measures 
the difference between two distributions. Let $\mathcal{P(\boldsymbol{\omega_{t/1}})}$, $\mathcal{P(\boldsymbol{\omega_{g}})}$ and $\mathcal{P(\boldsymbol{\omega_{t/k}})}$ denote the distributions corresponding to $\boldsymbol{\omega_{t/1}}$, $\boldsymbol{\omega_{g}}$ (global model weights) and $\boldsymbol{\omega_{t/k}}$, respectively.
In a learning process, we have:
    $D_{KL}(\mathcal{P(\boldsymbol{\omega}_{t/1})}, \mathcal{P(\boldsymbol{\omega}_{g}})) < D_{KL}(\mathcal{P(\boldsymbol{\omega}_{t/k})}, \mathcal{P(\boldsymbol{\omega}_{g})})$,
where $D_{KL}(P(X),Q(X))$ refers to the KLD between distribution $P(X)$ and $Q(X)$. According to the definition of KLD and the fact that $C_k$ has a skewed data distribution, we have (derivation details see~\ref{supp:derivation_eqcomp} in Supplement):
\vspace{-0.5em}
\begin{equation}\label{eq:comp}
  -\sum_{i=1}^{C}\mathcal{P^i(\boldsymbol{\omega_{t/1}})}\log(\mathcal{P^i(\boldsymbol{\omega_{g}})}) < -\sum_{i=1}^{C}\mathcal{P^i(\boldsymbol{\omega_{t/k}})}\log(\mathcal{P^i(\boldsymbol{\omega_{g}})}).   
\end{equation}
\vspace{-1.5em}

Eq.~\ref{eq:comp} can also be written as $\mathcal{L}(\boldsymbol{\omega}_{t/1})< \mathcal{L}(\boldsymbol{\omega}_{t_{/k}})$.  Recall Eq.~\ref{eq:inf}, it indicates $Inf(C_k)<Inf(C_1)$. As the round $t$ increases, combining $D_{KL}(\mathcal{P(\boldsymbol{\omega_{t/1}})}$, $\mathcal{P(\boldsymbol{\omega_{g}})}) \approx$ $D_{KL}(\mathcal{P(\boldsymbol{\omega_{t/k}})}$, $\mathcal{P(\boldsymbol{\omega_{g}})})$ as well as $ \sum_{i=1}^{C}\mathcal{P^i(\boldsymbol{\omega_{t/1}})}\log(\mathcal{P^i(\boldsymbol{\omega_{t/1}})}) \approx$ $\sum_{i=1}^{C}\mathcal{P^i(\boldsymbol{\omega_{t/k}})}\log(\mathcal{P^i(\boldsymbol{\omega_{t/k}})})$ gives $\mathcal{L}(\boldsymbol{\omega_{t_{/1}}}) \approx \mathcal{L}(\boldsymbol{\omega_{t_{/k}}})$. It follows that the Influence Index of $C_k$ and $C_1$ are approximately equal.

 \begin{proposition}
     \noindent\textit{The Influence Index is unfair towards clients with a skewed distribution in the initial phase of the training. During the later  phase of the training, it increases but remains unfair towards clients with larger-than-average amounts of data.}
 \end{proposition}

\textbf{Maverick - Size and Distribution Matters}
As by Eq.~\ref{eq:maverick}, a Maverick's data distribution differs considerably from an expected distribution and hence counts as a skewed distribution. Thus, the results for skewed data distributions hold for Mavericks. If a Maverick owns more data than an average client, it is treated doubly unfair as by the first evaluation regarding data size.

 \begin{proposition}
    \noindent \textit{ Mavericks are treated unfairly, with their Influence Index being lower than that of an average client in the early stage of the learning process and being similar in the latter stage,  despite their large quantity of data}. 
    \label{insight:3}
 \end{proposition}

\vspace{-1em}
\subsection{\SV for Mavericks}

\begin{definition}[\SV]
Let $\mathcal{K}$ denote the set of clients selected in a round excluding $C_k$, $\mathcal{K}\setminus\{C_k\}$ denote moving $C_k$ from $\mathcal{K}$. 
\SV of $C_k$ is 
\vspace{-0.5em}
\begin{equation}\label{eq:sv}
        SV(C_k) = \sum_{S \subseteq \mathcal{K} \setminus \{C_k\}} \frac{|S|!(|\mathcal{K}|-|S|-1)!}{|\mathcal{K}|!}\delta C_k(\mathcal{S}). 
\end{equation}
\end{definition}
\vspace{-1em}
Note that $\delta C_k(S)$ is the Influence Index on $S \cup C_k$, which is defined in Eq.~\ref{eq:inf}. To substitute the definition of influence index into the \SV~ in Eq.\ref{eq:sv}, we first analyze the difference of $C_k$ and $C_1$'s \SV, similar to Sec.~\ref{subsec:Influence}, resulting in (detailed derivation see \ref{supp:derivation_shapley} in Supplement):

\vspace{-2em}
\begin{equation}\label{eq:shapleyF}
    \begin{aligned}
SV(C_k)-SV(C_1)&=  \frac{1}{|\mathcal{K}|!}\bigg( (|\mathcal{K}|-1)! (\mathcal{L}(C_k)-\mathcal{L}(C_1))\\
    &+ \sum_{S \subseteq S_-} |S|!(|\mathcal{K}|-|S|-1)!(Inf_{S}(C_k)-Inf_{S}(C_1)) 
    + \sum_{S \subseteq S_+} |S|!(|\mathcal{K}|-|S|-1)!(Inf_{S}(C_k)-Inf_{S}(C_1))\bigg),
    \end{aligned}
\end{equation}
with $S_- = \mathcal{K} \setminus \{C_1, C_k\}$, $S_+ = \mathcal{K} \setminus \{C_1, C_k\} \cup{C_M}$, $C_M \in\{C_k,C_1\}$.

It can be concluded that the measurement of \SV and Influence Index share the same trend that they are unfair to Mavericks.
Note that rather than considering Influence Index for the complete set of $K$ clients, Eq.~\ref{eq:shapleyF}
only considers Influence Index on a subset $S$.
However, our derivations in Sec.~\ref{subsec:Influence}
are independent from the number of selected clients and remain applicable for subsets $S$, meaning that indeed the second and the third term of Eq.~\ref{eq:shapleyF} are negative. 
Similarly, the first term is negative as the loss for clients only owning one class is higher. 
However, \SV has better quality for clients with large data set than Influence Index since $\mathcal{L}(C_k)-\mathcal{L}(C_1)$ increases if the distance between $C_k$'s distribution and the global distribution is small. So, \iid clients with a large data quantity are evaluated better by \SV than Influence Index, in line with previous work. 

\begin{figure}[h]
	\centering
	{
	\subfloat[FM-1-excl. Maverick]{
	    \label{subfig:fashion-maverick1-shapley}
	    \includegraphics[width=0.24\textwidth]{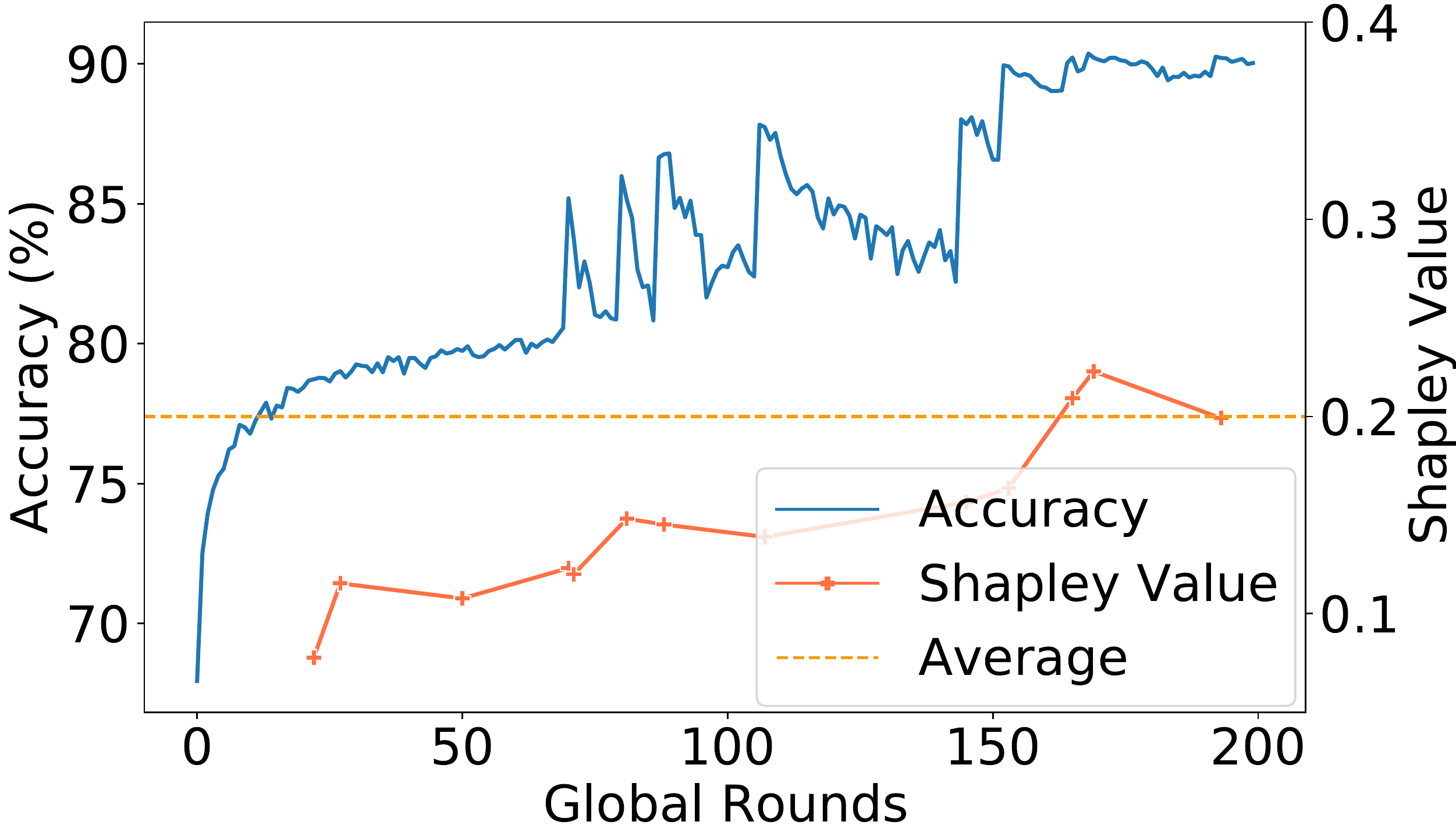} }
	\subfloat[FM-2-shared Mavericks]{
	    \label{subfig:fashion-maverick2M-shapley}
	    \includegraphics[width=0.24\textwidth]{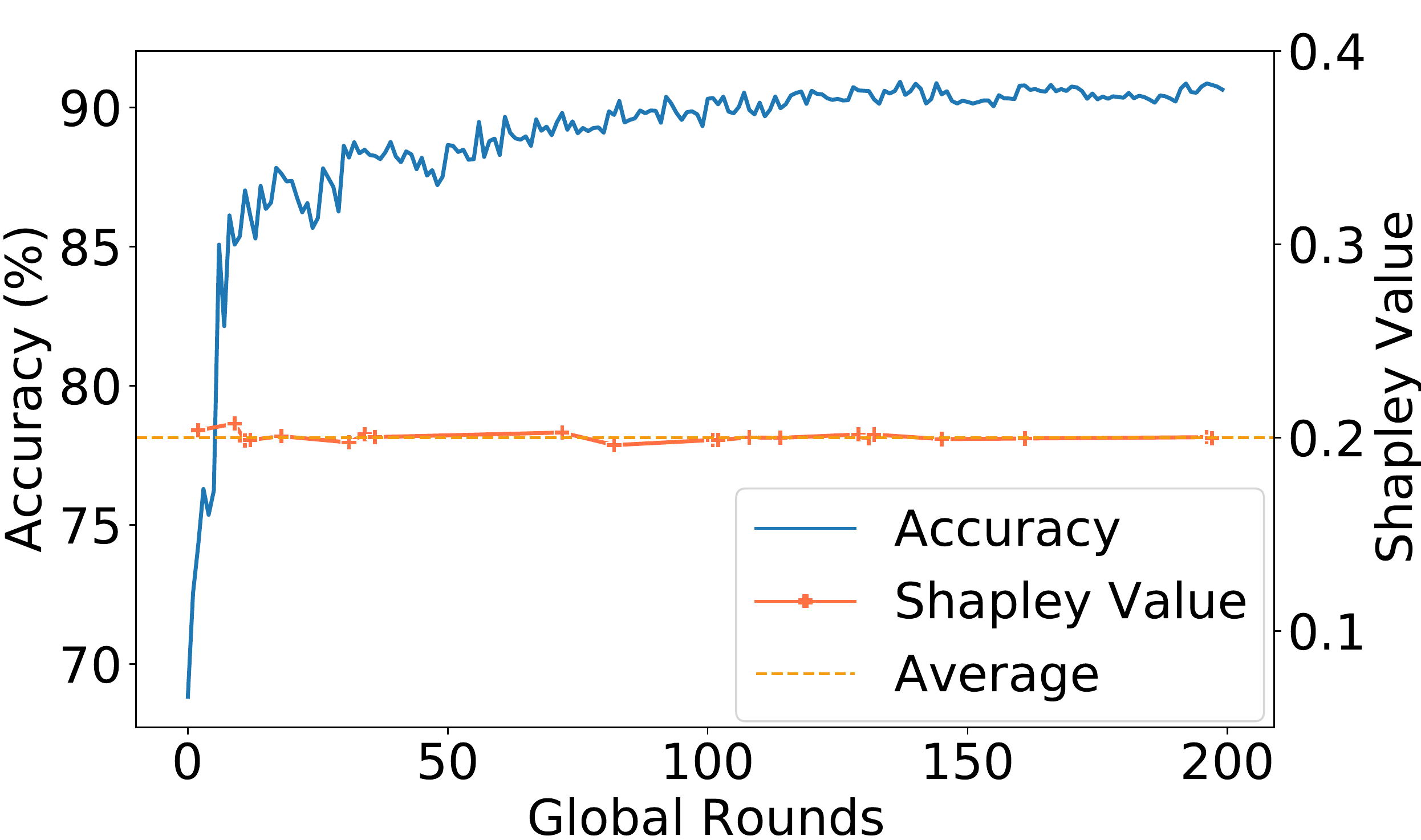} }
    \subfloat[FM-3-shared Mavericks]{
	    \label{subfig:fashion-maverick3M-shapley}
	    \includegraphics[width=0.24\textwidth]{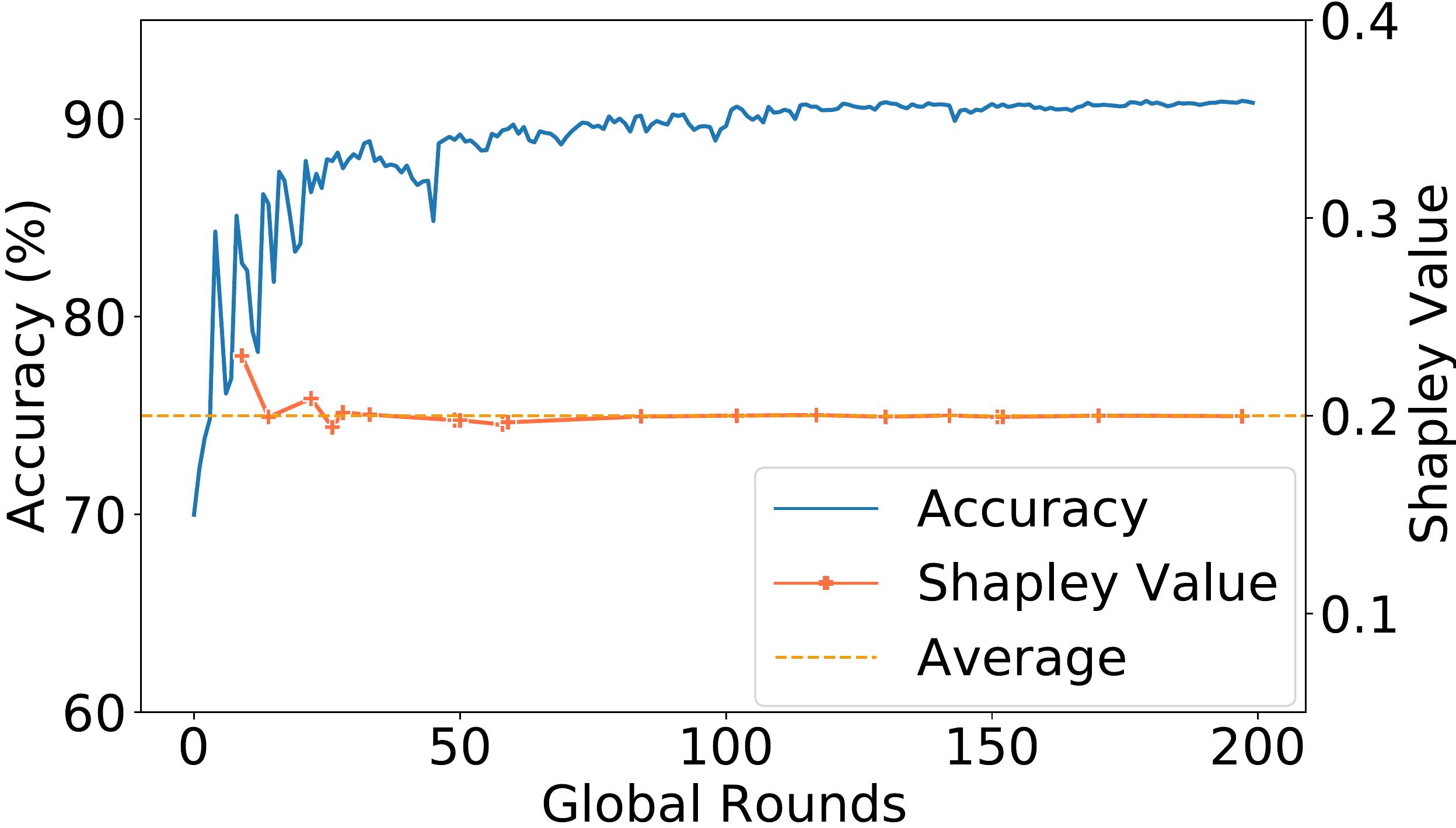}}
	\hfill
		\subfloat[C10-1-excl. Maverick]{
	    \label{subfig:cifar-maverick1-shapley}
	    \includegraphics[width=0.24\textwidth]{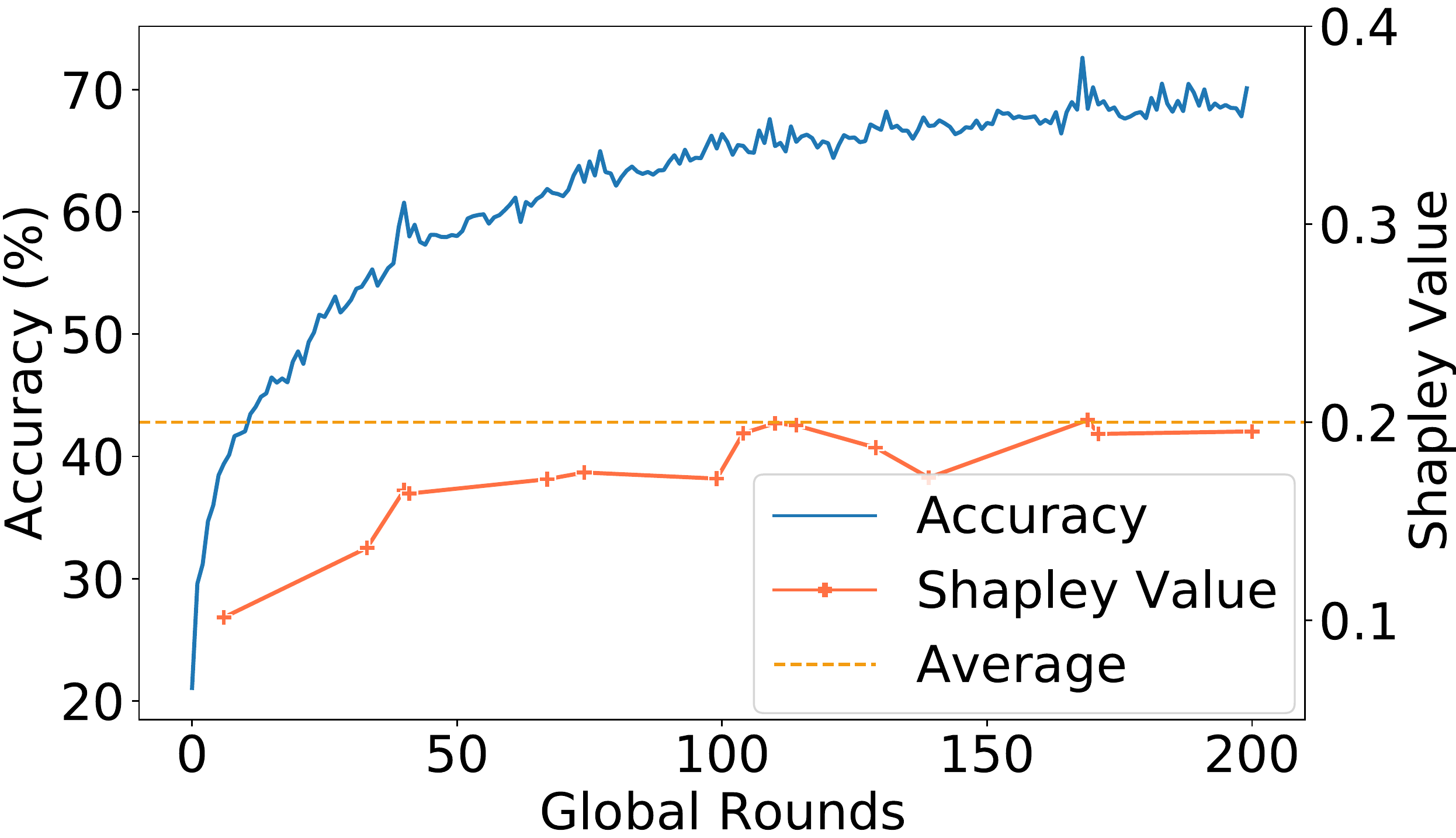}}
	\subfloat[FM-2-excl. Mavericks]{
	    \label{subfig:fashion-maverick2-shapley}
	    \includegraphics[width=0.24\textwidth]{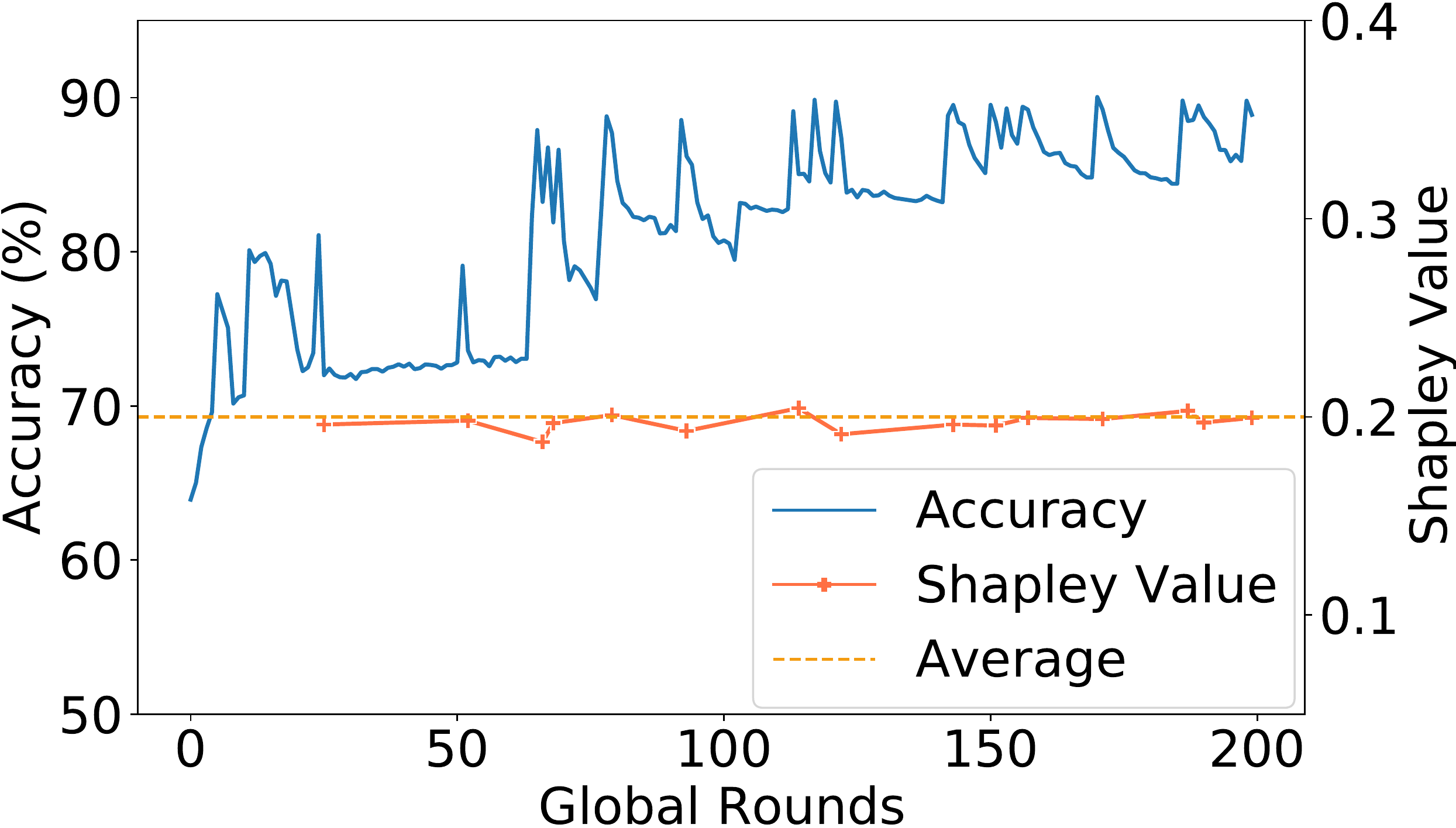} }
	\subfloat[FM-3-excl. Mavericks]{
	    \label{subfig:fashion-maverick3-shapley}
	    \includegraphics[width=0.24\textwidth]{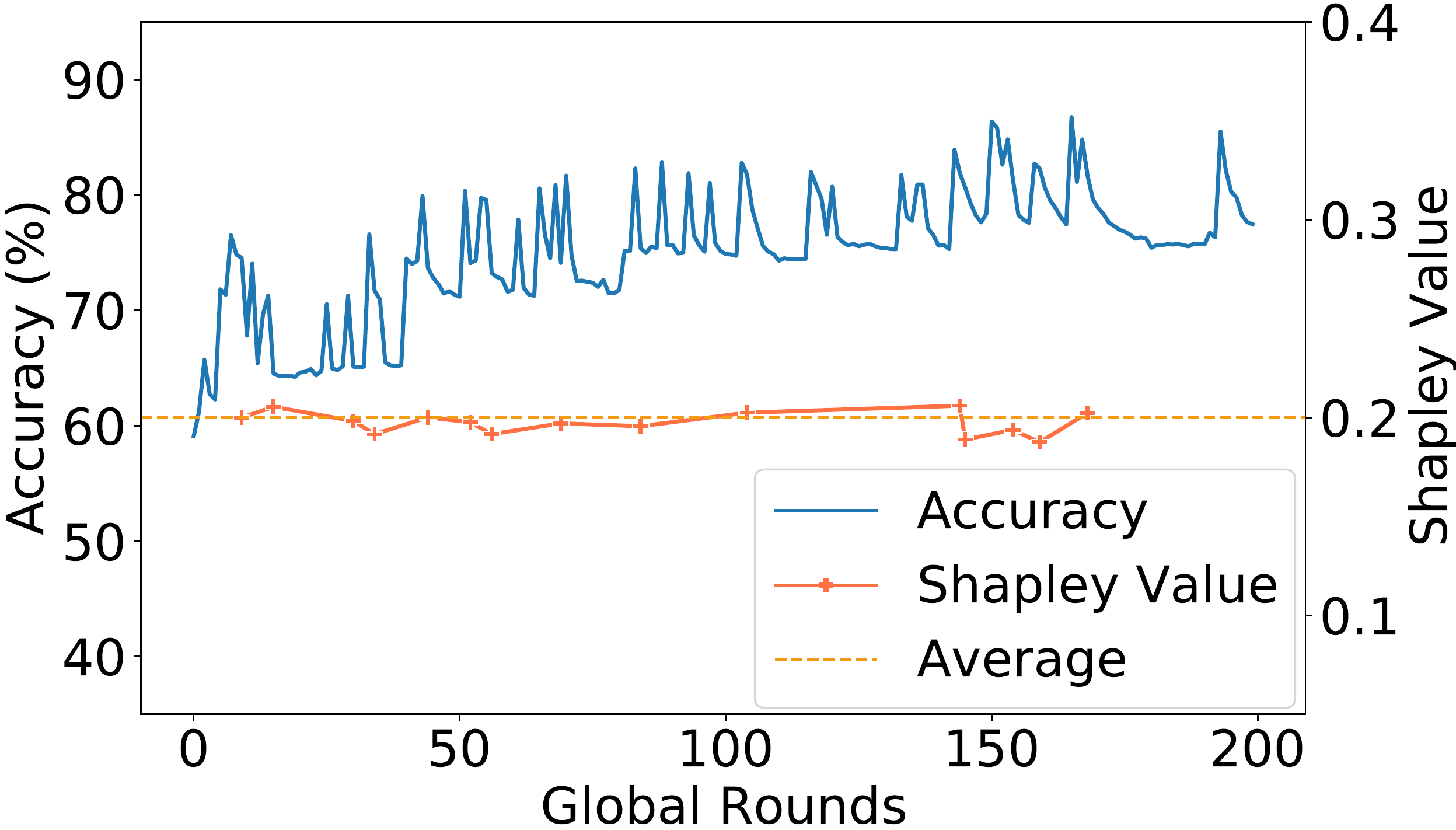}}
	}
	\caption{Relative \SV during training under multiple exclusive and shared Mavericks. }
	\label{fig:sv_eval}
\end{figure}

\textbf{Empirical Verification:}
We present the empirical evidences of how one or multiple Mavericks impact the global accuracy for an \SV-based contribution measure.
We consider 1,2, or 3 Mavericks. If more than 1 Maverick is considered, both exclusive and shared Mavericks are evaluated.
Hence, a Maverick owning a single class has considerably more data than non-Mavericks.
We use Fashion-MNIST (\ref{subfig:fashion-maverick1-shapley}) and Cifar-10~(\ref{subfig:cifar-maverick1-shapley}) as learning scenarios, with details in Sec.~\ref{sec:evaluation}.


Fig.~\ref{fig:sv_eval} shows the global accuracy and the relative \SV during the training, with the average relative \SV of the 5 selected clients indicated by the dotted line.
The contribution is only evaluated when a Maverick was selected. 
Fig.(\ref{subfig:fashion-maverick1-shapley}),
(\ref{subfig:cifar-maverick1-shapley}) confirm 
Proposition \ref{insight:3}.
For more Mavericks, the measured contribution is close to the average, as indicated in Fig.(\ref{subfig:fashion-maverick2M-shapley}),(\ref{subfig:fashion-maverick3M-shapley}),(\ref{subfig:fashion-maverick2-shapley}), and(\ref{subfig:fashion-maverick3-shapley}). Looking at the global accuracy for all scenarios, we see a fast increase with Mavericks, meaning that the initial phase during which the Maverick's contribution is lower than average is essentially non-existent. As even shared Mavericks own more data than non-Mavericks, an average \SV remains unfair.



%% file: samples/Sections/selection.tex
\section{\alg}
\label{sec:selection}
Here, we propose the 
novel client selection algorithm \alg, which enables FL systems with Mavericks to achieve faster convergence. First, we have established a trivial solution to empirically show that involving Mavericks in every round does not result in better accuracy. Quite the contrary, always including a Maverick has a detrimental effect in the long run.(See \ref{supp:emp_clientselection} in Supplement).
Further, we design \alg based on the Wasserstein Distance (EMD)~\cite{arjovsky2017wasserstein} of client data distributions as a light-weight dynamic client selection strategy. 

\SetKwInput{KwInput}{Input}                
\SetKwInput{KwOutput}{Output}  
\vspace{-0.2cm}

\makeatletter
\newcommand{\AlgoResetCount}{\renewcommand{\@ResetCounterIfNeeded}{\setcounter{AlgoLine}{0}}}
\newcommand{\AlgoNoResetCount}{\renewcommand{\@ResetCounterIfNeeded}{}}
\newcounter{AlgoSavedLineCount}
\newcommand{\AlgoSaveLineCount}{\setcounter{AlgoSavedLineCount}{\value{AlgoLine}}}
\newcommand{\AlgoRestoreLineCount}{\setcounter{AlgoLine}{\value{AlgoSavedLineCount}}}
\makeatother
\newcommand\mycommfont[1]{\footnotesize\ttfamily\textcolor{black}{#1}}
\SetCommentSty{mycommfont}

\textbf{Overview}. 
\alg (summarized in Alg.~\ref{alg:cs}) selects clients each round using weighted random sampling, where the weights of each client are dynamically updated every round. To obtain the weights, we compute the global and current Wasserstein distance for each client. We structure \alg in three steps : 
\textbf{\textit{i) Distribution Reporting and Initialization}} (Line 2--7): Clients perform distribution profile reporting so that the \federator is able to sum up the global distribution and initialize the current distribution. 
\textbf{\textit{ii) Dynamic Weights Calculation}} (Line 10--18): In this key step, we utilize a light-weight measure based on EMD to calculate dynamic selection probabilities over time, which achieve faster convergence, yet avoid overfitting. \textbf{\textit{iii) Weighted Client Selection }}(Line 19--23): According to the probabilities computed in the previous step, \alg applies weighted random sampling to select $K$ clients.

\textbf{Motivation of choice of Wasserstein Distance: }
 \alg considers both global and current distance on client data distribution to increase the chance of selecting Mavericks at the early stages and profit from their diverse data. Later on, the selection probability is reduced to avoid skewing the distribution towards the Maverick classes.
To measure the distance of distributions, the weight divergence defined by~\cite{zhao:2018:corr:noniid} of $\boldsymbol{\omega_t}$ and $\boldsymbol{\omega_t^k}$ is:
\begin{equation}\label{eq:wdif}
    \begin{aligned}
    \|\boldsymbol{\omega_t} - \boldsymbol{\omega_t^k}\|
    &=\eta \sum_{c=1}^{C}\|\sum_{k=1}^{K}\frac{n^i}{\sum_{i=1}^{K}}P^{i}(y=c)-P^k(y=c)\|\nabla_{\boldsymbol{\omega}}\mathbb{E}_{\boldsymbol{x} \mid y=c}\left[\log f_{c}(\boldsymbol{x}, \boldsymbol{\omega_{t-1}})\right].
    \end{aligned}
\end{equation}

Note that $\|\sum_{k=1}^{K}\frac{n^i}{\sum_{i=1}^{K}}P^{i}(y=c)-P^k(y=c)\|$ is the EMD of $C_k$ and the global distribution, which directly leads to the drift of $\boldsymbol{\omega_t}$ by $C_k$. Thus, we use EMD as the distance measure in our distribution-based \alg (Design principle and details see \ref{sec:design} in Supplement). 

\begin{wrapfigure}{R}{0.52\textwidth}
\vspace{-1em}
\hfill
\begin{minipage}{0.48\textwidth}
\begin{algorithm}[H]
    \label{alg:cs}
    \caption{\alg Client Selection}
    \KwData{$\mathcal{D}^i$ for $i \in {1, 2, ..., N}$.}
    \KwResult{
    $\mathcal{K}$: selected participants.
    }
     \textbf{Set:} distance coefficient $\alpha > 0$, $\beta >0$;\\
  initialize probability $Proba^1$;\\
  initialize current distribution $\mathcal{D}_c^{1}$;\\
    $\mathcal{D}_g \leftarrow \sum_{i=1}^N \mathcal{D}^i$;\\
    \For{client i = 1, ..., N}
        {
        $emd_g.append(EMD(\mathcal{D}_g, \mathcal{D}^i))$;\\
        $\widetilde{emd}_{g} \leftarrow Norm(emd_g, \mathcal{D})$
        }
    \For{round t = 1, 2, ..., R}
        {
    initialize output list $\mathcal{K}_t= range(K)$;\\
    \For{client i = 1, ..., K}{
        $u_i \leftarrow rand(0,1)$; $k_i \leftarrow u_i(\frac{1}{Proba^t_i})$;\\
        \For{client j = K+1, K+2, ..., N}{
            $T \leftarrow \min(k)$, $k = \{k_1, k_2, ...,k_i\}$;
            $k_j \leftarrow u_j(\frac{1}{Proba^t_j})$;\\
        \eIf{$k_j> T$:}
          {$\mathcal{K}_t[\min(k).index] \leftarrow j$}
             {continue}}
             }
        $\mathcal{D}_c^{t+1} \leftarrow \mathcal{D}_c^{t} + \sum_{s \in \mathcal{K}} \mathcal{D}^s$;\\
        \For{client i = 1, ..., N}
        {
            $emd_c.append(EMD(\mathcal{D}_c^t, \mathcal{D}^i))$;\\
            $\widetilde{emd}_{c} \leftarrow Norm(emd_c, \mathcal{D})$
        }
        \For{client i = 1, ..., N}
        {        
    $Proba^{t+1}_i \leftarrow softmax(\alpha \widetilde{emd}_{g}[i] - t \beta \widetilde{emd}_{c}[i])$
        }
    }
\end{algorithm}
\end{minipage}
\vspace{-2em}
\end{wrapfigure}

\textbf{Dynamic Weights Calculation:}
Let $EMD(\mathcal{P_1}, \mathcal{P_2})$ be the EMD of distributions between $\mathcal{P_1}$ and $\mathcal{P_2}$ and $\mathcal{D}$ be the self-reported data distribution matrix of $N$ clients. It is only reported once at the beginning of the learning. 
Also, we assume a limit of $R$ on the global communication rounds. The dynamic selection probability of clients in round $t$, $Proba_t$, is based on both the global distance $emd_{g}$ and current distance $emd_{c}$ between the local distribution $\mathcal{D}_i$ and the global/current $\mathcal{D}_g$, $\mathcal{D}_c^t$ (Line 23). Assume that  $c_1$ is one class randomly chosen by \federator except for the Maverick class\footnote{the \federator can tell from the reported data distributions which classes are Maverick classes} from $\mathcal{D}$, here we apply normalization method $Norm(P, \mathcal{D}) = \frac{\sum_{c=1}^CP(y=c)}{\sum_{i=1}^NP^i(y=c_1)/N}$ for distributions $P$ (Line 7, 21). The current distribution is the accumulated of selected clients over rounds (Line 18).

\textbf{Distance-based Aggregation:} The larger $\widetilde{emd}_{g}$ is, the higher the probability $Proba_i^{t}$ that a client $C_i$ is selected, since $C_i$ brings more distribution information to train $\boldsymbol{\omega_t}$. However, the current distance is also taken into consideration to improve the selection of Mavericks.
Therefore, $Proba^t_i = softmax(\alpha \widetilde{emd}_{g}[i] -t \beta  \widetilde{emd}_{c}[i])$ for $C_i$ in round $t$, where $\alpha, \beta$ is the distance coefficient to weigh the global and current distance. $\alpha, \beta$ shall be adapted for different initial distributions, i.e., different dataset and distribution rules (See \ref{supp:parameter} in Supplement). Clients with larger global distance and smaller current distance have a high probability to be selected, and thus Mavericks are involved more often during the early period and less later on of training to gain faster convergence. The sampling procedure for selecting $K$ clients based on $Proba^t$ (Line 9--17) has complexity of  $O(K\log(\frac{N}{K}))$~\cite{efraimidis:2006:IPL:ARes}, so comparably low.

%% file: samples/Sections/evaluation.tex
\section{Experimental Evaluation}
\label{sec:evaluation}
\vspace{-1em}
In this section, we evaluate the convergence speed of \alg in comparison to four state-of-the-art selection strategies and FedProx as the representative of SOTA algorithm for data heterogeneity.
The evaluation considers four different classifiers and two Maverick types. 


\textbf{Datasets and Classifier Networks }  We use public image datasets: \textit{i)} Fashion-MNIST~\cite{xiao:2017:online:fashion} for bi-level image classification; \textit{ii)} MNIST~\cite{lecun:1998:mnist} for simpler tasks which need less data to do fast learning; \textit{iii)} Cifar-10~\cite{krizhevsky:2009:cifar} for more complex task such as colored image classification; \textit{iv)} STL-10~\cite{coates:2011:stl} for applications with small amounts of local data for all clients. 
We note that light-weight neural networks are more applicable for federated learning scenarios, where clients typically have limited computation and communication resources~\cite{muhammad:2020:KDD:fedfast}. Thus, here we apply light-weight CNN for each dataset correspondingly, neural network details are provided in~\ref{supp:datasets} in Supplement.

\textbf{Federated Learning System } The system considered has 50  participants with homogeneous computation and communication resources and 1 \federator. At each round, the \federator selects 5 (a common practise in related studies~\cite{tolpegin2020data}) clients according to a client selection algorithm.
The \federator uses FedSGD or FedAvg to aggregate local models from selected clients.
As FedSGD always considers one local epoch, 
we choose the same for FedAvg to enable a fair comparison of the two approaches. 
Two types of Mavericks are considered: exclusive and shared Mavericks. We consider the case of single Maverick owning an entire class of data in most of our experiments.  

\textbf{Evaluation Metrics } \textit{i)} Global test accuracy for all classes; \textit{ii)} Source recall (related results see~\ref{supp:emp_clientselection} in Supplement) for classes owned by Mavericks exclusively;
 \textit{iii)} $R@99$:  the number of communication rounds required to reach 99\% of test accuracy of random selection based results; 
 \textit{iv)} Normalized \SV ranging between $[0, 1]$ to measure the contribution of Mavericks; 5) Computation time (results see~\ref{supp:overhead} in Supplement): the computation time measured by the \federator, including training, aggregation and client selection time.

\textbf{Baseline } We consider four selection strategies: \textbf{Random}~\cite{mcmahan:2017:aistat:fedavg}, \textbf{\SV-based},  \textbf{FedFast}~\cite{muhammad:2020:KDD:fedfast}, and recent \textbf{TiFL}~\cite{Zheng:2020:HPDC:Tier}\footnote{When implementing, we focus on their client selection and leave out other features. For example, we do not implement the features related to the communication acceleration in TiFL and the aggregation in FedFast.} under both FedSGD and FedAvg aggregation methods.  Although \SV itself is a contribution measure rather than client selection strategy, it has been widely adopted in incentive design to attract the participants of clients with high \SV~\cite{sim2020collaborative}.  Thus, the \SV-based algorithm here is a variant of weighted random selection whose weights are decided by the relative \SV, i.e., the probability of a client to be selection corresponds to their relative \SV. Further, in order to compare with 
state-of-the-art solution for heterogeneous FL which focus on the optimization design, we evaluate \textbf{FedProx}~\cite{Tian:2020:mlsys:Heterogeneous} as one of the baselines.

\vspace{-1em}
\subsection{Convergence Analysis for Selection}
\begin{wrapfigure}{r}{9cm}	
	\vspace{-2em}
	{
	\subfloat[FMNIST-FedSGD]{
	    \label{subfig:fashion-sgd}
	    \includegraphics[width=0.3\textwidth]{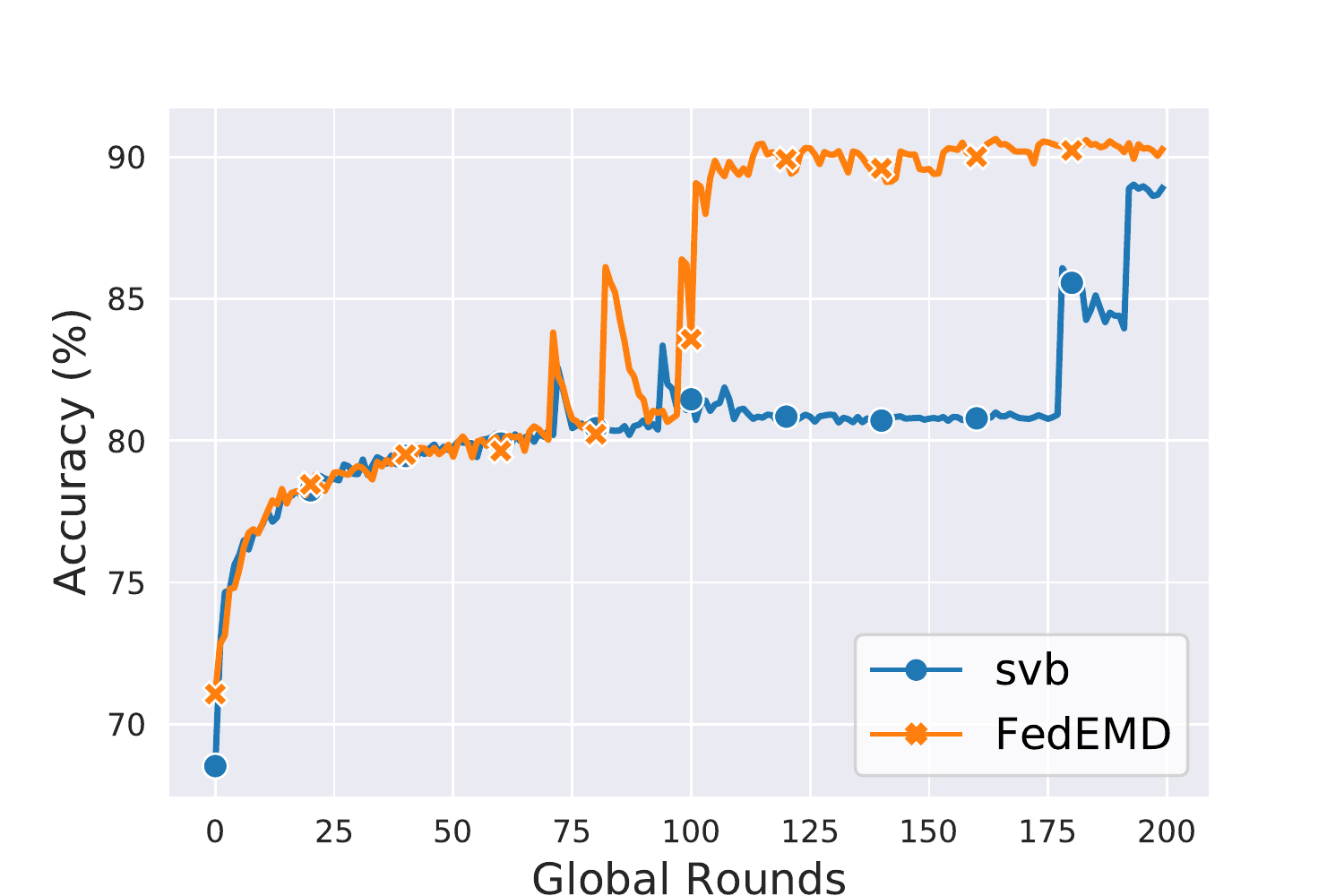} }
	\subfloat[FMNIST-FedAvg]{
	    \label{subfig:fashion-avg}
	    \includegraphics[width=0.3\textwidth]{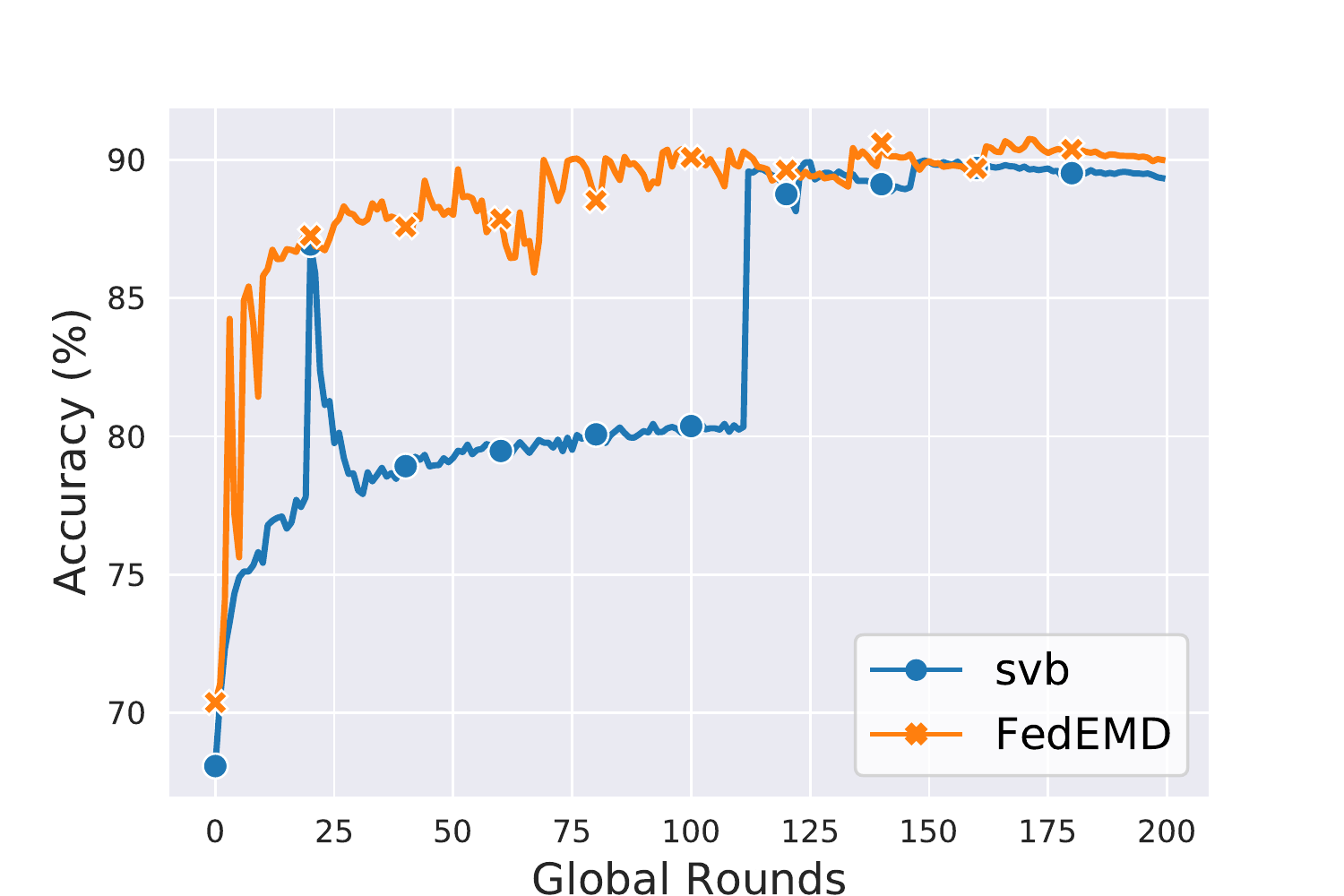}}
	}
	\caption{Comparison on \alg with \svb.}
	\label{img:new_vs_sv}
	\vspace{-1em}
\end{wrapfigure}
Fig.~\ref{img:new_vs_sv} shows the change in global accuracy over rounds. \alg achieves an accuracy close to the maximum almost immediately for FedAvg while \svb requires about 100 rounds. For FedSGD, both client selection methods have a slower convergence but \alg still only requires about half the number of times to achieve the same high accuracy as \svb. The reason for the result lies in \svb rarely selecting the Maverick in the early phase of the training, as, by Sec.~\ref{sec:shapley}, the Maverick has a below-average \SV. 

Furthermore, let us zoom into the number of rounds needed to reach 99\% of max accuracy. Combining these results with random selection, we see that it takes 72 and 104 rounds to reach $R@99$ for \svb and \alg in FedAvg, respectively while \svb fails in reaching $R@99$ within 200 rounds. 

\textbf{Comparison with baselines}. We summarize the comparison with the state-of-the-art methodologies in Table~\ref{tab:convergence_speed}. The reported $R@99$ is averaged over three replications. Note that we run each for 200 rounds, which is mostly enough to see the convergence statistics for these lightweight networks. The rare exceptions when 99\% maximal accuracy is not achieved for random selection are indicated by  
$>200$, e.g., when FedFast fails in most of evaluations for Maverick.

Thanks to its distance-based weights, \alg achieves faster convergence than all other algorithms consistently. The reason for such result is that \alg enhances the participation of the Maverick during early training period, speeding up learning global distribution.
For most settings, the difference in convergence speed is considerably visible. 
\begin{table}[H]
\setlength\tabcolsep{1pt}
\centering
\caption{Convergence speed of client selection strategies in $R@99$ Accuracy.}
\label{tab:convergence_speed}
\begin{tabular}{c|ccccc|ccccc} 
\toprule

& \multicolumn{5}{c|}{FedSGD}                                                          & \multicolumn{5}{c}{FedAvg}    \\
\small
\multirow{-2}{*}{\textbf{Dataset} } & \textbf{Random}
 & \textbf{TiFL} & \textbf{FedFast} &  \textbf{FedProx}& \textbf{\alg}  & \textbf{Random} & \textbf{TiFL} & \textbf{FedFast} & \textbf{FedProx} & \textbf{\alg}   \\ 
  \midrule
\textit{MNIST}                                                                                   &            132.7              &          111.0           &           >200             &   117.7  & \underline{\textbf{98.7}}                 &       72.3                   &         84.0            &          >200             &  51.0   &\underline{\textbf{40.0}}              \\ 
\textit{FMNIST}                                                                                        &             144.0             &           140.3          &           >200             & 135.7 &     \underline{\textbf{131.3}}       &                    110.7             &          146.3           &           >200            &  92.0    & \underline{\textbf{79.7}}                \\ 
\textit{Cifar-10}                                                                                        &           140.7               &          147.3           &             >200          &   164.0  & \underline{\textbf{140.0}}        &                      143.0          &         119.7            &          173.7             &  143.7   &  \underline{\textbf{107.0}}              \\ 
\textit{STL-10}                                                                                      &            122.3              &          124.7           &          171.0             &  186.0    & \underline{\textbf{96.3}}       &                   179.7               &           >200          &           153.0            &   179.0   & \underline{\textbf{95.0}}                 \\
  \bottomrule
  \end{tabular}
\end{table}

The only exception here are relatively easy tasks with simple averaging rather than weighted, e.g., Cifar-10 with FedSGD, which indicates our distribution-based client selection method is especially useful for data size-aware aggregation and more complex tasks (e.g., FedAvg). 
While such an increased weight caused by larger data size can lead to a decrease in accuracy in the latter phase of training, Mavericks are rarely selected in the latter phase of training by \alg, which successfully mitigates the effect and achieves a faster convergence. 

\subsection{Different Types of Mavericks}
We explore the effectiveness of \alg on two types of Mavericks: exclusive and shared Mavericks.
\begin{wrapfigure}{r}{9.5cm}
	\vspace{-2.5em}
	\centering
	{
 	\subfloat[Exclusive Mavericks]{
 	    \label{subfig:R70multiple}
 	    \includegraphics[width=0.32\textwidth]{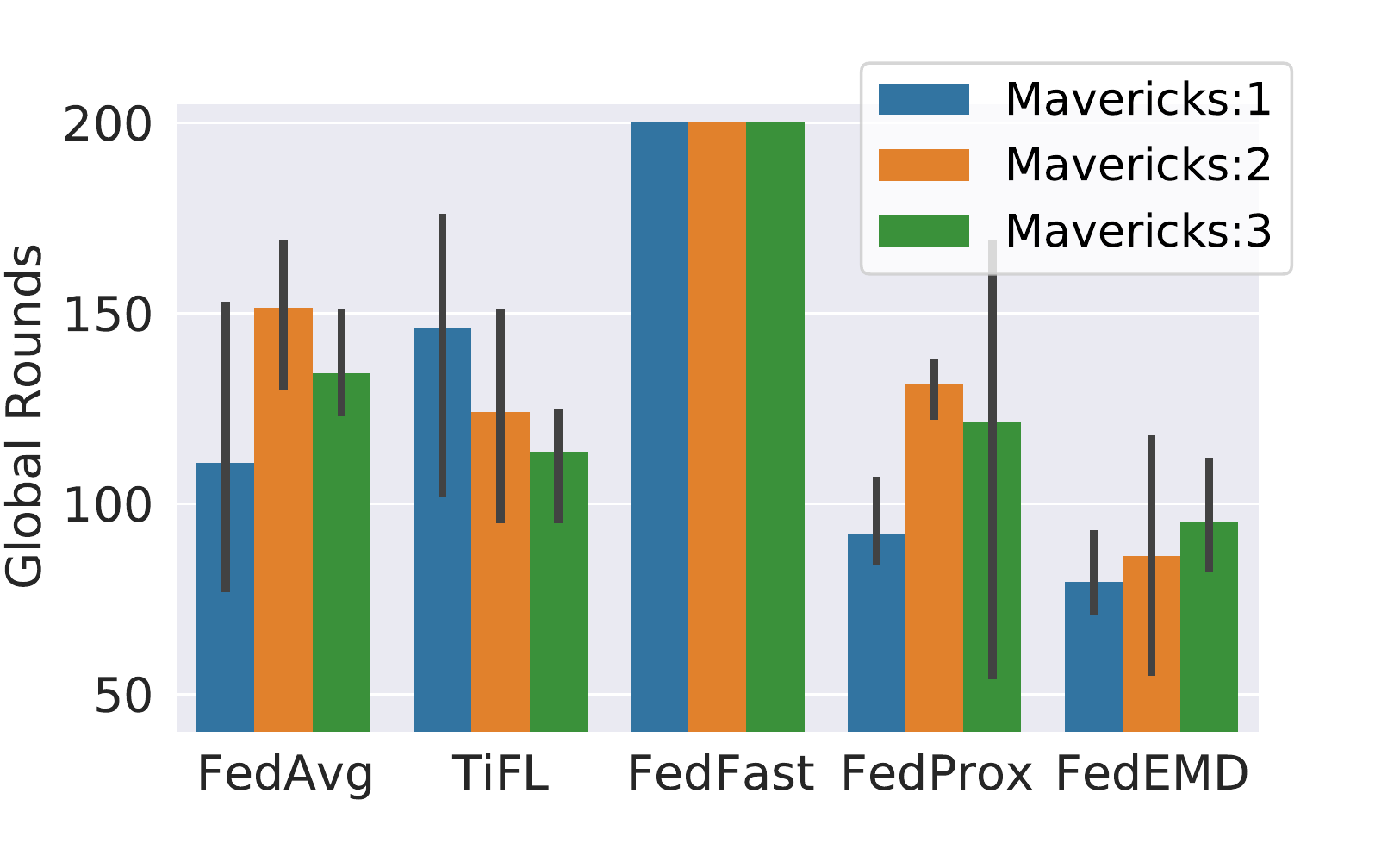}}
    \subfloat[Shared Mavericks]{
 	    \label{subfig:R90multiple}
	    \includegraphics[width=0.32\textwidth]{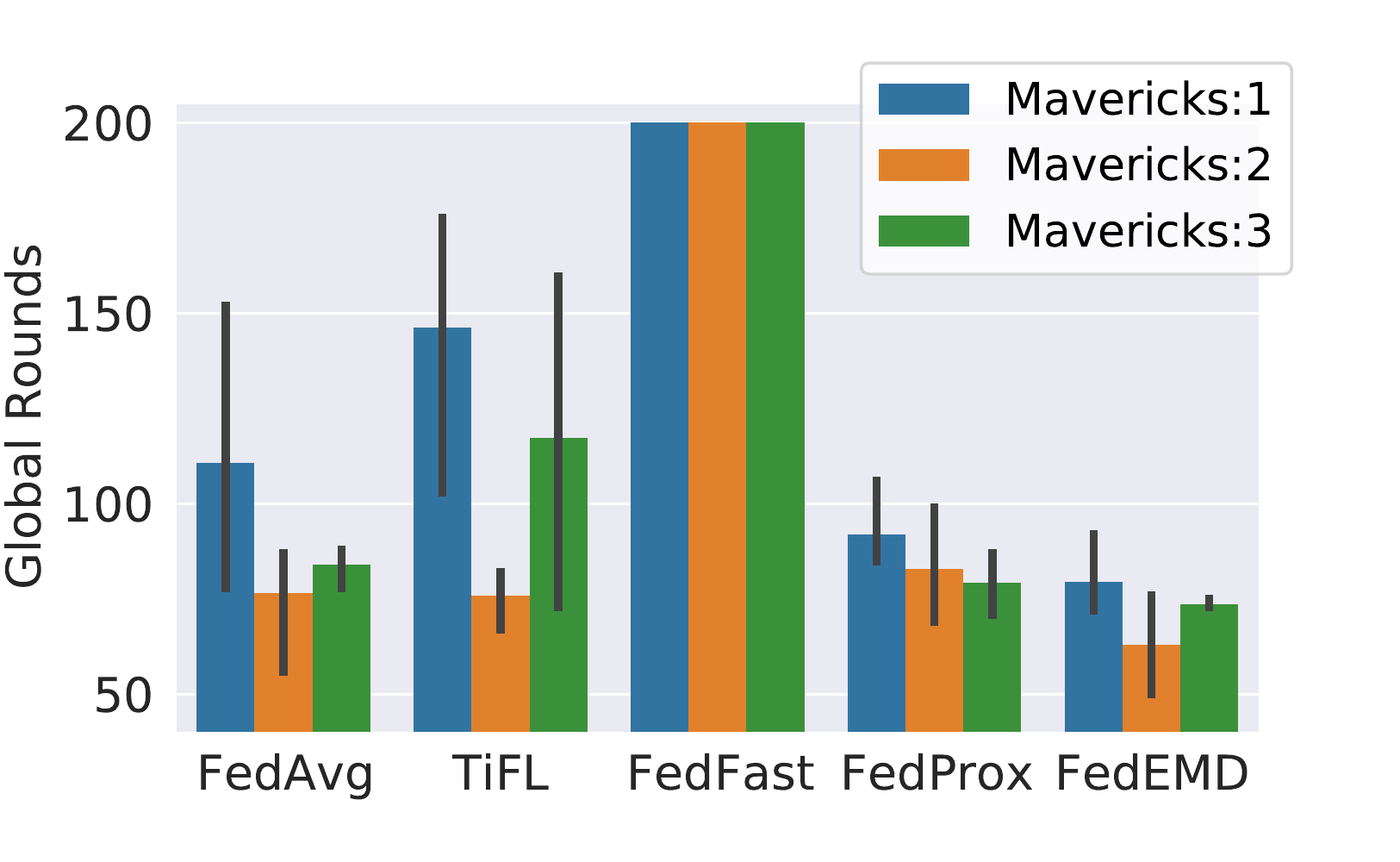}}
	}
	\caption{Convergence speed $R@99$ for multiple Mavericks.}
	\label{img:multiple}
	\vspace{-1.5em}
\end{wrapfigure}
We vary the number of Mavericks between one and three and use the Fashion-MNIST dataset. The Maverick classes are `T-shirt', `Trouser', and `Pullover'. Results are shown with respect to $R@99$. 

Fig.~\ref{subfig:R70multiple} illustrates the case of multiple exclusive Mavericks. For exclusive Mavericks, the data distribution becomes more skewed as more classes are exclusively owned by Mavericks. 
\alg always achieves the fastest convergence, though its convergence time increases slightly as the number of Mavericks increases, reflecting the increased difficulty of learning in the presence of skewed data distribution.  
FedFast's $K$-mean clustering typically results in a cluster of Mavericks and then always includes at least one Maverick. As shown in~\ref{sec:design} in Supplement, constantly including a Maverick hinders convergence, which is also reflected in FedFast's results. 
TiFL outperforms FedAvg with random selection for multiple Mavericks. However, TiFL's results differ drastically over runs due to the random factor in local testing. Thus, TiFL is not a reliable choice for Mavericks. Comparably, FedProx tends to achieve the best performance among the SOTA algorithms but still exhibits slower convergence than \alg because higher weight divergence entails higher penalty on the loss function.

For shared Mavericks, a higher number of Mavericks indicates a more balanced distribution. Similar to the exclusive case, \alg has the fastest convergence and FedFast again trails the others. The improvement of \alg over the other methods is less visible due to the limited advantage of \alg on balanced data. In terms of the effectiveness of \alg handing more shared Mavericks, the convergence speed improves slightly. However, we attribute such an observation partially to the fact that a higher number of Mavericks resembles the case of \iid. Random performs the most similar to \alg, as random selection is best for \iid scenarios, which shared Mavericks are closer to. Note that the standard deviation of \alg is smaller, implying a better stability. 



%% file: samples/Sections/conclusion.tex
\section{Conclusion}
\label{sec:conclusion}

Client selection is key to successful Federated Learning as it 
enables maximizing the usefulness of different diverse data sets.
In this paper, we highlighted that existing schemes fail when clients have heterogeneous data, in particular if one class is exclusively owned by one or multiple Mavericks. 
We proposed and evaluated an alternative algorithm that encourages the selection of diverse clients at the opportune moment of the training process, accelerating the convergence speed by up to at least 26.9\% in average for multiple dataset under FedAvg aggregation, compared to the state of the art.

\clearpage



%% file: samples/Sections/supplement.tex
\section{Supplement}

\subsection{The Derivation of Eq.~\ref{eq:shapleyF}}
\label{supp:derivation_shapley}
From Eq. \ref{eq:sv}, for a set $S\subseteq \mathcal{K}$,  $S_{+C_k} = S \cup \{C_k\}$, and we write $\delta C_k(S) = \mathcal{L}(S_{+C_k}) -\mathcal{L}(S)$ where $\mathcal{L}(S_{+C_k}) = \mathcal{L}(\boldsymbol{\omega}^{(S_{+C_k})})$
We analyze the difference of $C_k$ and $C_1$'s \SV, similar to Sec.~\ref{subsec:Influence}, resulting in: 

\vspace{-0.5cm}
\begin{equation}\label{eq:svdif1}
    \begin{aligned}
    SV(C_k)-SV(C_1) &= \frac{1}{|\mathcal{K}|!} \left(\sum_{S \subseteq \mathcal{K} \setminus \{C_k\}} |S|!(|\mathcal{K}|-|S|-1)! \delta C_k(S)\right.\\
    &\left.- \sum_{S \subseteq \mathcal{K} \setminus \{C_1\}} |S|!(|\mathcal{K}|-|S|-1)!\delta C_1(S)\right).
    \end{aligned}
\end{equation}

Recall that  $\delta C_k(S) = \mathcal{L}(S_{+C_k}) -\mathcal{L}(S) = Inf(C_k)$,  we will have:
Eq.~\ref{eq:svdif1} can be written as:
\vspace{-0.2cm}
\begin{equation}\label{eq:svdif}
    \begin{aligned}
    SV(C_k)-SV(C_1)   &= \frac{1}{|\mathcal{K}|!}\bigg( (|\mathcal{K}|-1)! (\mathcal{L}(C_k)-\mathcal{L}(C_0))\\
    &+ \sum_{S \subseteq S_-, S\neq \emptyset} |S|!(|\mathcal{K}|-|S|-1)!(Inf_{S}(C_k)-Inf_{S}(C_1)) \\
    &+ \sum_{S \subseteq S_-} (|S|+1)!(|\mathcal{K}|-|S|-2)!Inf_{S}(C_k) \\
    &-\sum_{S \subseteq S_-} (|S|+1)!(|\mathcal{K}|-|S|-2)!Inf_{S}(C_1)\bigg),
    \end{aligned}
\end{equation}
where $S_- = \mathcal{K} \setminus \{C_1, C_k\}$. The first term of the four term corresponds to the case when $S = \emptyset$.  The second term treats the subsets in which neither $C_k$ nor $C_1$ are contained. The third and the fourth term treat the cases when $C_1$ is included for the computation of $SV(C_k)$ and $C_k$ is included in the computation of $SV(C_1)$, respectively. The inclusion of another element is mirrored by considering $|S|+1$ instead of $|S|$. 
Eq.~\ref{eq:svdif} corresponds to:  
\vspace{-0.1cm}
\begin{equation}\label{eq:shapley_final}
    \begin{aligned}
    SV(C_k)-SV(C_1)&=  \frac{1}{|\mathcal{K}|!}\bigg( (|\mathcal{K}|-1)! (\mathcal{L}(C_k)-\mathcal{L}(C_1))\\
    &+ \sum_{S \subseteq S_-} |S|!(|\mathcal{K}|-|S|-1)!(Inf_{S}(C_k)-Inf_{S}(C_1)) \\
    &+ \sum_{S \subseteq S_+} |S|!(|\mathcal{K}|-|S|-1)!(Inf_{S}(C_k)-Inf_{S}(C_1))\bigg),
    \end{aligned}
\end{equation}
with $S_- = \mathcal{K} \setminus \{C_1, C_k\}$, $S_+ = \mathcal{K} \setminus \{C_1, C_k\} \cup{C_M}$, $C_M \in\{C_k,C_1\}$.

\subsection{The Derivation of Eq.~\ref{eq:comp}}
\label{supp:derivation_eqcomp}
To derive:

\begin{equation}
    \begin{aligned}
    D_{KL}(\mathcal{P(\boldsymbol{\omega}_{t/1})}, \mathcal{P(\boldsymbol{\omega}_{g}})) < D_{KL}(\mathcal{P(\boldsymbol{\omega}_{t/k})}, \mathcal{P(\boldsymbol{\omega}_{g})})\,
    \end{aligned}
    \label{eq:dkl_ineq}
\end{equation}
we start the derivation from Eq.~\ref{eq:dkl_ineq}. The left side of the inequality \ref{eq:dkl_ineq} equals

\begin{equation}\label{eq:kl1_supp}
    \sum_{i=1}^{C}\mathcal{P^i(\boldsymbol{\omega_{t/1}})}\log(\mathcal{P^i(\boldsymbol{\omega_{t/1})}}) - \sum_{i=1}^{C}\mathcal{P^i(\boldsymbol{\omega_{t/1}})}\log(\mathcal{P^i(\boldsymbol{\omega_{g}})})
\end{equation}
while the right side can be written as:

\begin{equation}\label{eq:kl2_supp}
    \sum_{i=1}^{C}\mathcal{P^i(\boldsymbol{\omega}_{t/k})}\log(\mathcal{P^i(\boldsymbol{\omega}_{t/k})}) - \sum_{i=1}^{C}\mathcal{P^i(\boldsymbol{\omega}_{t/k})}\log(\mathcal{P^i(\boldsymbol{\omega}_{g})}).
\end{equation}

Note that $\sum_i P(x_i)\log(P(x_i))$ is the entropy of distribution $P(X)$. If $C_k$ has a skewed distribution, $\mathcal{P(\boldsymbol{\omega}_{t/1})}$ has larger entropy than $\mathcal{P(\boldsymbol{\omega}_{t/k})}$ as $\mathcal{P(\boldsymbol{\omega}_{t/1})}$ includes the skewed distribution of $C_k$.  Hence, at an early stage of the training, it holds that

\begin{equation}\label{eq:entropy_supp}
     \sum_{i=1}^{C}\mathcal{P^i(\boldsymbol{\omega_{t/1}})}\log(\mathcal{P^i(\boldsymbol{\omega_{t/1}})}) > \sum_{i=1}^{C}\mathcal{P^i(\boldsymbol{\omega_{t/k}})}\log(\mathcal{P^i(\boldsymbol{\omega_{t/k}})})
\end{equation}

Based on Eqs.~\ref{eq:kl1_supp}, \ref{eq:kl2_supp}, and Inequality \ref{eq:entropy_supp}, we have: 

\begin{equation*}
  -\sum_{i=1}^{C}\mathcal{P^i(\boldsymbol{\omega_{t/1}})}\log(\mathcal{P^i(\boldsymbol{\omega_{g}})}) < -\sum_{i=1}^{C}\mathcal{P^i(\boldsymbol{\omega_{t/k}})}\log(\mathcal{P^i(\boldsymbol{\omega_{g}})}).   
\end{equation*}


\subsection{Empirical Verification of Data Size Fairness}
\label{supp:emp_fairness}

\begin{figure}[h]
	\centering
	{
 	\subfloat[Global fairness utility for Fashion-MNIST]{
 	    \label{subfig:utility-fashion}
 	    \includegraphics[width=0.5\textwidth]{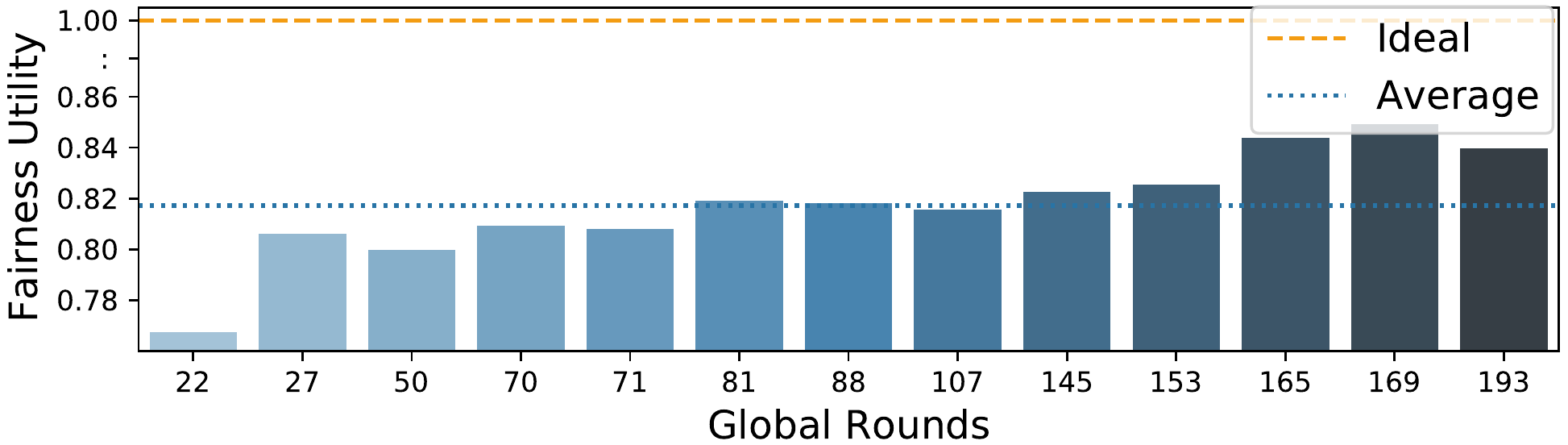}}
	\hfill
    \subfloat[Global fairness utility for Cifar-10]{
	    \label{subfig:utility-cifar}
	    \includegraphics[width=0.5\textwidth]{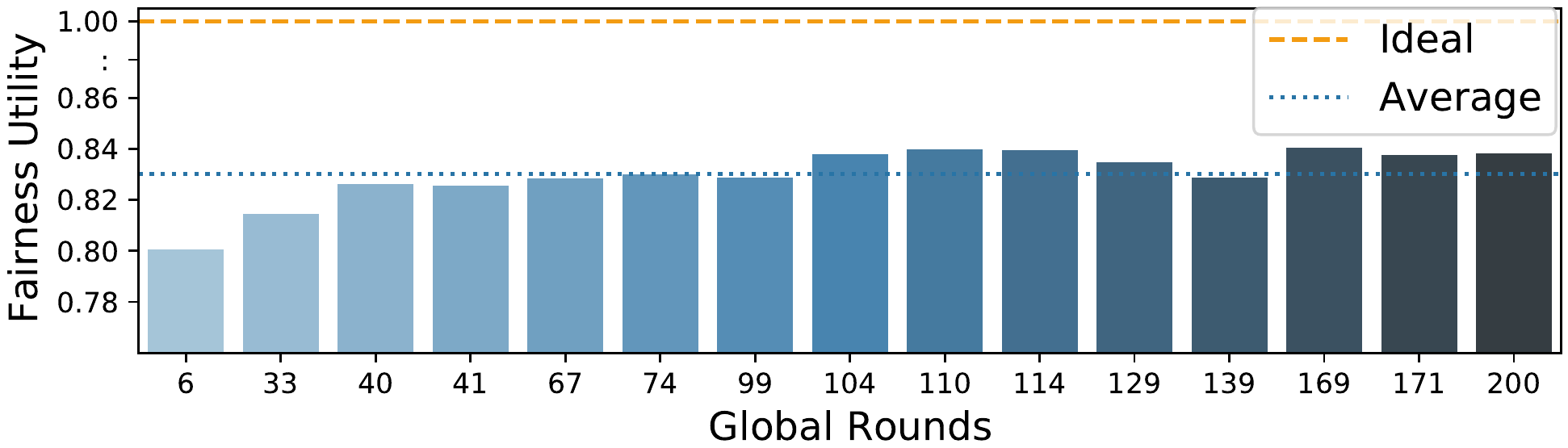}}
	}
	\caption{Data size fairness measured by \SV with one exclusive Maverick.}
	\label{fig:utility}
\end{figure}

We illustrate the aspect of fairness in Fig.~\ref{fig:utility}, which is defined by Def.~\ref{eq:fairnessClient}. In line with our analysis, fairness improves slightly over time but remains low in the presence of Mavericks. Concretely, the system-wide fairness increases from slightly below 0.8 to close to 0.85, with an ideal situation being 1.0. The result is consistent for both data sets.

\subsection{Empirical Observation of Client Selection}
\label{supp:emp_clientselection}

We evaluate a simple selection protocol that knows the identity of the one Maverick in the system and always selects them as one of the clients. Our example evaluation uses Fashion-MNIST dataset, other corresponding settings in detail are involved in Section \ref{sec:evaluation}. The one Maverick owns class `Trouser' exclusively. For each class to be trained on the same number of data points, the Maverick owns $K$ times as much data as the other clients. All clients have equal communication and computation resources. Our main metrics are the global accuracy and the accuracy on class ‘Trouser’, which depends solely on the Maverick. We provide upper and lower bounds on the performance by comparing our setting with an \iid distribution over all classes (denoted by \iid) and one over all classes without the Maverick class (denoted by mav-never), respectively. When including a Maverick, we use random selection (denoted by mav-random) to compare with the case when the Maverick is always selected (denoted by mav-always).  

\begin{figure}[th]
\centering
\setlength{\abovecaptionskip}{-0cm}   
\includegraphics[width=0.8\textwidth]{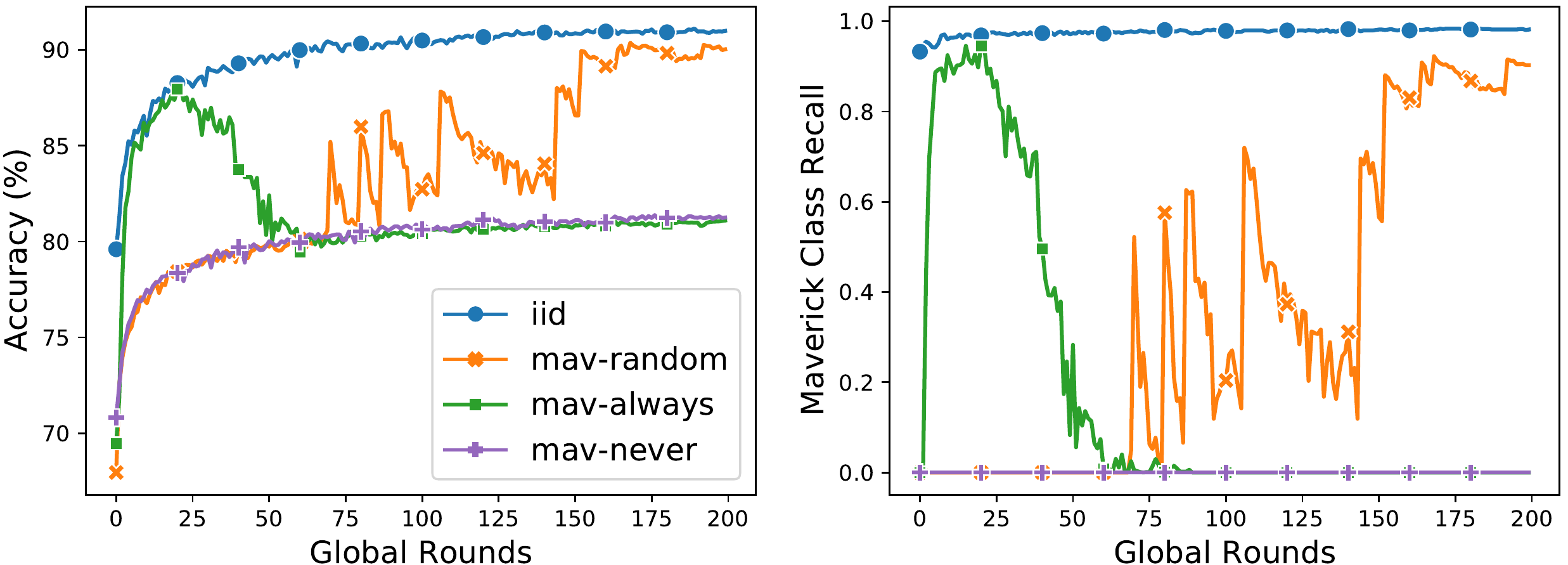}
\setlength{\abovecaptionskip}{0.1cm}
\caption{Different strategies of involving Maverick on Fashion-MNIST: (i) mav-never: never choosing Maverick, (ii)  mav-random: randomly choosing clients, and (iii) mav-always: always choosing Maverick. \iid is the reference performance. Left figure shows the global accuracy and the right figure shows the recall of Maverick class.}
\label{img:participation}
\end{figure}

Fig.~\ref{img:participation} indicates that while initially outperforming random selection and being similarly accurate as the upper bound of \iid data distribution, always including a Maverick has a detrimental effect in the long run. The result can be explained as the inclusion of Mavericks enlarges the distance between the global distribution and the learned distribution. 
Concretely, the learned distribution is skewed towards the Maverick class as the majority of images it trains on is from the Maverick class, with the Maverick having more data than all other selected clients combined. 

\subsection{Design Support for \alg}
\label{sec:design}
Based on our insights from the previous experiment (See Sec~\ref{supp:emp_clientselection} in Supplement), we design \alg to involve Mavericks in the beginning but avoid overtraining by involving them less later on. 
In this manner, we achieve faster convergence and a light-weight protocol in comparison to \SV. 
We utilize $A\%$-convergence as the global round when the test accuracy reaches $A\%$ of the highest accuracy reached when using random selection. 
 Here, we further assume that clients record their local data quantity per class to the \federator. It has been shown that such reporting is indeed acceptable for privacy-preserving~\cite{Zheng:2020:HPDC:Tier}.
 
 Three factors contributes to our selection method: i) Global distance, which is the distribution differences between single client and the aggregated global distribution of all clients. It indicates the differences from global distribution of each client and the one with higher global distance has high probability to be selected. ii) Current distance, which is the distance between the accumulated distribution over each rounds of the selected clients' distribution and a single client. A larger current distance indicates that selecting the client might deter convergence, hence the probability of the client to be selected should be lower. iii) Round decay, which increases the impact of current distance over time to avoid non-convergence.

Hence, \alg considers both global and current distance on client data distribution to bias the selection strategy to involve Mavericks at the early stage with higher probability to increase convergence speed, while reducing the selection probability of Mavericks once the round decay becomes relevant to avoid skewing the distribution towards the Maverick classes.
To measure the distance of distributions, the weight divergence~\cite{zhao:2018:corr:noniid} of $\boldsymbol{\omega_t}$ and $\boldsymbol{\omega_t^k}$ is:
\vspace{-0.2cm}
\begin{equation}\label{eq:wdif}
    \begin{aligned}
    \|\boldsymbol{\omega_t} - \boldsymbol{\omega_t^k}\|
    &= \|\sum_{i=1}^{K}\frac{n^i}{\sum_{i=1}^{K}}(\boldsymbol{\omega_{t-1}^{k}} - \eta \sum_{c=1}^{C}P^{i}(y=c)\nabla_{\boldsymbol{\omega}}\mathbb{E}_{\boldsymbol{x} \mid y=c}\left[\log f_{c}(\boldsymbol{x}, \boldsymbol{\omega_{t-1}})\right])\\
    &-\sum_{k=1}^{K}\frac{n^k}{\sum_{k=1}^{K}}\boldsymbol{\omega_{t-1}^{k}} + \eta \sum_{c=1}^{C}P^k(y=c)\nabla_{\boldsymbol{\omega}}\mathbb{E}_{\boldsymbol{x} \mid y=i}\left[\log f_{i}(\boldsymbol{x}, \boldsymbol{\omega_{t-1}})\right]\|\\
    &=\eta \sum_{k=1}^{K}\frac{n^k}{\sum_{k=1}^{K}}\sum_{i=1}^{C}\|P^{k}(y=i)\nabla_{\boldsymbol{\omega}}\mathbb{E}_{\boldsymbol{x} \mid y=i}\left[\log f_{i}(\boldsymbol{x}, \boldsymbol{\omega_{t-1}})\right]\\
    &-P(y=i)\nabla_{\boldsymbol{\omega}}\mathbb{E}_{\boldsymbol{x} \mid y=i}\left[\log f_{i}(\boldsymbol{x}, \boldsymbol{\omega_{t-1}})\right])\|_2\\
    &=\eta \sum_{c=1}^{C}\|\sum_{k=1}^{K}\frac{n^i}{\sum_{i=1}^{K}}P^{i}(y=c)-P^k(y=c)\|\nabla_{\boldsymbol{\omega}}\mathbb{E}_{\boldsymbol{x} \mid y=c}\left[\log f_{c}(\boldsymbol{x}, \boldsymbol{\omega_{t-1}})\right].
    \end{aligned}
\end{equation}
Note that $\|\sum_{k=1}^{K}\frac{n^i}{\sum_{i=1}^{K}}P^{i}(y=c)-P^k(y=c)\|$ is the Wasserstein Distance (EMD) of $C_k$ and the global distribution, which directly leads to the drift of $\boldsymbol{\omega_t}$ by $C_k$. Thus, we use Wasserstein Distance as the distance measure in our distribution-based \alg. 
 The Wasserstein Distance is defined as:
\begin{equation}\label{eq:EMD}
\operatorname{EMD}\left(P_{r}, P_{\theta}\right)=\inf _{\gamma \in \Pi} \sum_{x, y}\|x-y\| \gamma(x, y)=\inf _{\gamma \in \Pi} \mathbb{E}_{(x, y) \sim \gamma}\|x-y\|,
\end{equation}
where $\Pi (P_r, P_\theta)$ represents the set of all possible joint probability distributions of $P_{r}, P_{\theta}$. $\gamma(x,y)$ represents the probability that $x$ appears in $P_{r}$ and $y$ appears in $P_{\theta}$.

Additionally, compared with other popular distance measures, Wasserstein Distance is suitable in our case because: i) even if the support sets of the two distributions do not overlap or overlap very little, it still reflects the distance of two distributions, while the Jensen–Shannon Distance (JSD) \cite{endres:2003:TIT:JSD} is constant in this case and the KLD approaches $\infty$; ii) it is more smooth in the sense that it avoids sudden changes in distance, in contrast to JSD.

\subsection{Observation on the Impact of Mavericks}

 we first conduct experiments of learning a neural network classifier for Fashion-MNIST under scenarios of \iid and in the presence of one Maverick owning all images of class `Trouser'. We study the convergence over the global rounds, considering four scenarios: i) random selection\footnote{The default client selection in FL is random policy.} and \iid distribution with equal quantity (\iid-random), ii) \SV-based selection and 
\iid distribution with heterogeneous data quantity (size-het-svb), iii) random selection and one Maverick (maverick-random), and  iv) \SV-based selection and one Maverick (maverick-svb). 
\label{supp:hypothesis}
\begin{figure}[H]
\centering
\setlength{\abovecaptionskip}{-0cm}   
\includegraphics[width=0.5\textwidth]{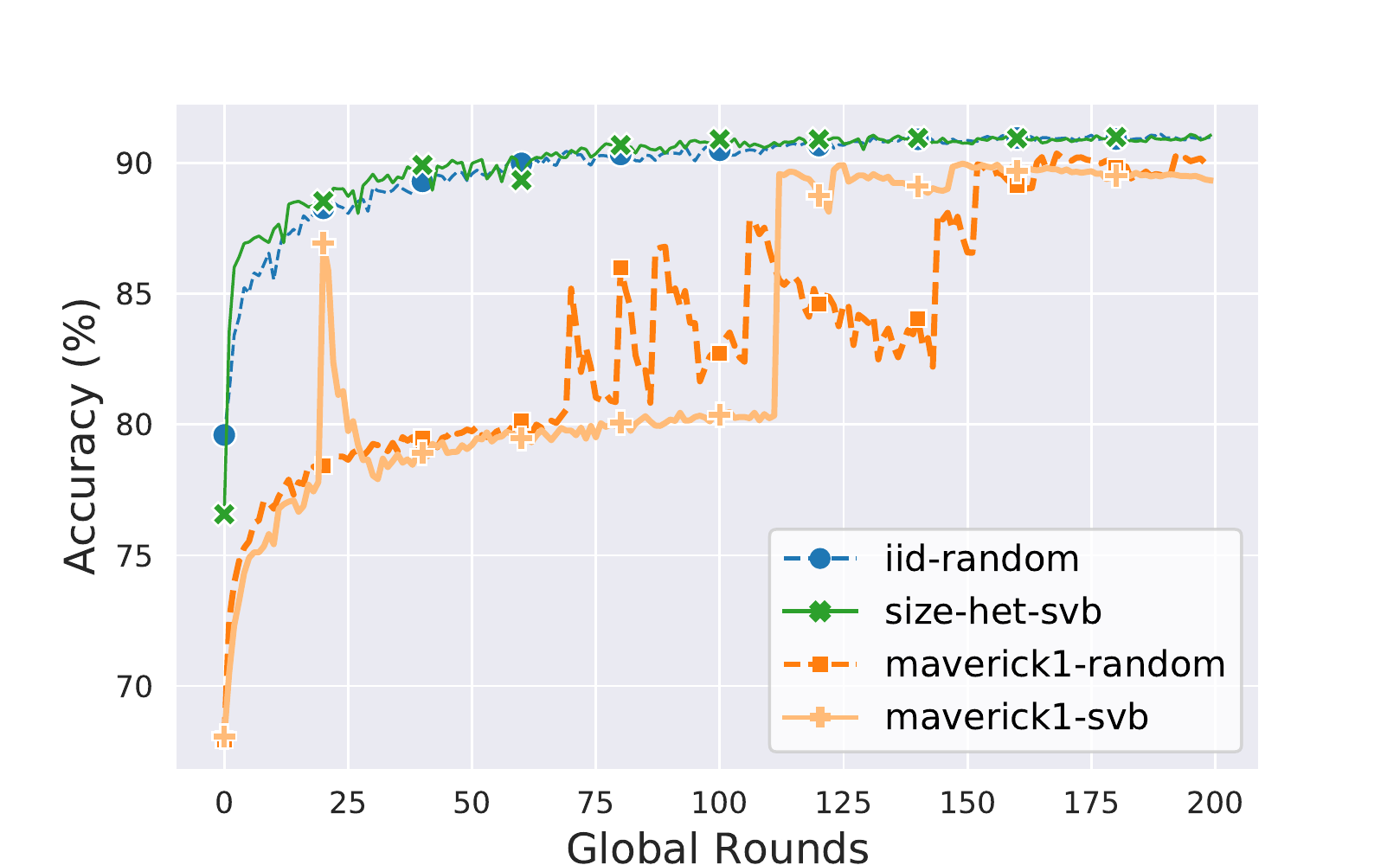}
\setlength{\abovecaptionskip}{0.1cm}
\caption{Comparing random and \SV based (\svb) selection of clients for \iid and in the presence of Maverick (Fashion-MNIST dataset)}
\label{img:rand_vs_sv}
\end{figure}

Figure~\ref{img:rand_vs_sv} shows that random and \SV-based selection both perform well in the scenarios that they are designed for. Yet, they both exhibit slow convergence for Mavericks, with \SV not showing a clear advantage over the random selection. This result leads us to conjecture that \SV is unable to fairly evaluate Mavericks' contribution. 

\subsection{Computational Overhead of \federator}
\label{supp:overhead}


\begin{table}[!h] 
\caption{Computational Overhead of \federator on Fashion-MNIST}
\centering
 \label{tab:time}

 \begin{tabular}{cccccc}  

\toprule   

  Algorithm  & FedAvg & TiFL & FedFast & FedProx & \alg \\

\midrule   

  Overhead & 1 & 2.4 & 1.76 & 1.56 & 1.04 \\

  \bottomrule  

\end{tabular}

\end{table}

Our overhead is measured by the computation time at the \federator, see Table~\ref{tab:time}, with all experiments run on the same machine. We express the computation time relative to the baseline of random selection.
On Fashion-MNIST dataset with FedAvg, 
the computation times for TiFL, FedFast, and \alg increase by a factor 2.4, 1.76, and 1.04. respectively, in comparison to random. As expected, \alg has the lowest overhead due to its light-weight design. TiFL and FedFast have notably higher overheads since TiFL performs expensive local model testing and FedFast performs K-means clustering for client selection each round and always choose Maverick whose training takes long. These results are consistent for all data sets, so we present only  Fashion-MNIST due to space constraints. 

\subsection{Hardware and Software}
\label{supp:hardware}
We develop such a FL emulator via Pytorch and run experiments on Ubuntu 20.04, with 32 GB memory which is equipped with a GeForce RTX 2080 Ti GPU and a Intel i9 CPUs with 10 cores (2 threads each).

\subsection{Details of the Datasets and Networks}
\label{supp:datasets}

The key characteristics of the datasets and corresponding neural networks are summarized in Table~\ref{tab:datasets}.
\begin{table}[H]
\centering
\caption{Statistics overview of dataset.}
\label{tab:datasets}
\begin{tabular}{|c|c|c|c|c|} 
\hline
\textbf{Dataset} & \textbf{Type} & \textbf{Train} & \textbf{Classes} & \textbf{Test} \\ 
\hline
\hline
\textit{MNIST}   & bilevel-28*28 & 60K & 10 & 10K \\ 
\hline
\textit{Fashion-MNIST}     & bilevel-28*28 & 60K & 10 & 10K \\ 
\hline
\textit{Cifar-10}   &  colored-28*28 & 50K & 10 & 10K \\ 
\hline
\textit{STL-10}   &  colored-96*96 & 5K & 10 & 8K \\
\hline
\end{tabular} \\
\end{table}

\textbf{MNIST:} For the image classification of handwritten digits in MNIST, a very simple CNN with 2-convolutional layer followed by a densely-connected layer is used which is able to achieve accuracy of 99.5\%. The output is a class label from 0-9.
\textbf{Fashion-MNIST:} For the image classification of bilevel fashion related images in Fashion-MNIST, the neural network is the same as CNN in MNIST.  \textbf{Cifar-10:} For this colored image classification task, the classes are completely mutually exclusive. Neural network applied is CNN with 6-convolutional layer and 2 densely-connected layer to map the output from 0-9. \textbf{STL-10:} This dataset consists of only 500 data in each class, in our federated setting, each non-Maverick owns 10 data in one label. It also use a shallow CNN with 3-convolutional layer followed by 2 fully-connected layer.

\subsection{Recommend Parameters}
\label{supp:parameter}
\begin{itemize}
    \item \emph{MNIST: }Batch size: 4; LR: 0.001; Momentum: 0.5; Scheduler step size: 50; Scheduler gamma: 0.1; $\alpha$: 0.15, $\beta:$ 0.0015;
    \item \emph{Fashion-MNIST: }Batch size: 4; LR: 0.001; Momentum: 0.9; Scheduler step size: 10; Scheduler gamma: 0.1; $\alpha$: 0.15, $\beta:$ 0.0015;
    \item \emph{Cifar-10: }Batch size: 10; LR: 0.01; Momentum: 0.5; Scheduler step size: 50; Scheduler gamma: 0.5; $\alpha$: 0.11, $\beta:$ 0.0015;
    \item \emph{STL-10: }Batch size: 10; LR: 0.01; Momentum: 0.5; Scheduler step size: 50;  Scheduler gamma: 0.5; $\alpha$: 0.11, $\beta:$ 0.001;
\end{itemize}

\section{Broader Impact}

\subsection*{Generalization and Limitations}
\label{sec:limitation}

We consider the contribution measurement and client selection in the presence of Mavericks, who hold large data quantities and exhibit skewed data distributions. When the number of exclusive Mavericks increases to the extreme (i.e. all over the clients are Mavericks), the Maverick scenario approaches the data heterogeneous scenarios considered in  prior work~\cite{zhao:2018:corr:noniid}. When the number of shared Mavericks increases, the FL system approaches an \iid scenario. 
In this paper, we do not consider differences in computational or network resources. We suggest to combine \alg with prior work ~\cite{nishio:2019:ICC:FedCS,huang:2020:tpds:RBCS-F} to avoid selecting clients with insufficient resources. 

\subsection*{Discrimination in Federated Learning}
We have shown that Federated Learning discriminates against clients with unusual data both when assigning rewards and during the selection procedure. 
Note that this is in line with work on security enhancements for FL, which have also been shown to exclude users with diverse data~\cite{wang2020attack}.
Such a bias against clients with diverse data is likely to exclude people from minority groups from the learning process, as they can have data that is very specific to their experience as a member of a minority group. As a consequence, minority groups might be excluded from profiting from the learning outcome. 
Our difference enables to promote clients with diverse data strategically and is hence a first step to increasing diversity in Federated Learning. However, the profit of this step is limited, as long as security measures and lacking incentives are still likely to be biased against certain groups. More work is needed to ensure that Federated Learning does not amplify already existing social and institutional discrimination.